\documentclass[sigconf,natbib=true]{acmart}
\usepackage{amsmath,amsfonts}
\usepackage{adjustbox}

\usepackage{appendix}
\usepackage{algorithm}
\usepackage[noend]{algpseudocode}
\usepackage{graphicx}
\usepackage{textcomp}
\usepackage{xcolor}
\usepackage{caption}
\usepackage{wrapfig}
\usepackage{subcaption}
\usepackage{tabularx}
\usepackage{multirow}
\usepackage{makecell}
\usepackage{diagbox}
\usepackage{booktabs}
\usepackage{color, colortbl}
\usepackage{courier}
\usepackage{pifont}
\definecolor{Gray}{gray}{0.9}
\algrenewcommand\algorithmicrequire{\textbf{Input:}}
\algrenewcommand\algorithmicensure{\textbf{Output:}}

\AtBeginDocument{
  \providecommand\BibTeX{{
    \normalfont B\kern-0.5em{\scshape i\kern-0.25em b}\kern-0.8em\TeX}}}

\setcopyright{acmcopyright}
\copyrightyear{2024}
\acmYear{2024}

\author{Yoonhyuk Choi}
\orcid{0000-0003-4359-5596}
\affiliation{
  \institution{
  Seoul National University}
  \city{Seoul}
  \state{Republic of Korea}
}
\email{younhyuk95 @ snu.ac.kr}

\author{Jiho Choi}
\orcid{0000-0002-7140-7962}
\affiliation{
  \institution{
  Seoul National University}
  \city{Seoul}
  \state{Republic of Korea}
}
\email{jihochoi @ snu.ac.kr}

\author{Taewook Ko}
\orcid{0000-0001-7248-4751}
\affiliation{
  \institution{
  Seoul National University}
  \city{Seoul}
  \state{Republic of Korea}
}
\email{taewook.ko @ snu.ac.kr}

\author{Chong-Kwon Kim}
\orcid{0000-0002-9492-6546}
\affiliation{
  \institution{Korea Institute of Energy Technology}
  \city{Naju}
  \state{Republic of Korea}
}
\email{ckim @ kentech.ac.kr}

\renewcommand{\shortauthors}{Yoonhyuk Choi et al.}

\newtheorem{manualtheoreminner}{\textbf{Theorem}}
\newenvironment{manualtheorem}[1]{%
  \renewcommand\themanualtheoreminner{#1}%
  \manualtheoreminner
}{\endmanualtheoreminner}

\newtheorem{manualcorollary}{\textbf{Corollary}}
\newenvironment{manualcoroll}[1]{%
  \renewcommand\themanualcorollary{#1}%
  \manualcorollary
}{\endmanualcorollary}

\newtheorem{manuallemma}{\textbf{Lemma}}
\newenvironment{manuallem}[1]{%
  \renewcommand\themanuallemma{#1}%
  \manuallemma
}{\endmanuallemma}

\acmConference[Conference '25] {under review}{}{}
\acmBooktitle{under review}
\acmPrice{15.00}
\acmISBN{978-1-4503-9236-5/22/10}
\acmDOI{10.1145/}

\begin{document}

\title{Improving Signed Propagation for Graph Neural Networks in Multi-Class Environments}



\begin{abstract}


Graph Neural Networks (GNNs) perform well on homophilic networks, where edges generally connect nodes with the same label.
However, their performance declines on heterophilic networks, prompting the investigation of alternative message-passing approaches.
Notably, assigning negative weights to heterophilic edges (signed propagation) has gained attention, with some studies confirming its effectiveness theoretically.
However, these studies often assume binary classification, which may not generalize to multi-class graphs.
To address this, we provide new theoretical insights into GNNs in multi-class settings, highlighting the limitations of signed propagation regarding both message-passing and parameter updates.
Our findings show that ignoring feature distribution in signed propagation can reduce the separability of dissimilar neighbors, increasing prediction uncertainty and instability.
To overcome these challenges, we propose two novel calibration strategies that enhance discrimination and reduce entropy in predictions.
Both theoretical and extensive experimental analyses demonstrate that our methods improve the performance of signed and general GNNs.

\end{abstract}



\maketitle

\section{Introduction}
The recent proliferation of graph-structured datasets has ignited rapid advancements in graph mining techniques. Graph Neural Networks (GNNs) provide satisfactory performances in various downstream tasks including node classification and link prediction. The main component of GNNs is message-passing \cite{defferrard2016convolutional}, where information is propagated between adjacent nodes and aggregated. Additionally, the entailment of a structural property enhances the representation and the discrimination powers of GNNs
\cite{kipf2016semi,velickovic2017graph}. 

Early GNNs assume the network homophily, where nodes of similar features make connections based on social influence and/or selection theories. Spectral GNNs \cite{defferrard2016convolutional,kipf2016semi} employ Laplacian smoothing called low-pass filtering to receive low-frequency signals from neighbor nodes. Consequently, these methods fail to adequately address heterophilous graphs \cite{zhu2020beyond} and show dismal performance \cite{ma2021homophily} in several heterophilous cases. To resolve this problem, numerous clever algorithms (spatial GNNs) have been proposed, including adjustment of edge coefficients \cite{velickovic2017graph,brody2021attentive} and aggregation of remote but highly similar nodes \cite{pei2020geom,li2022finding}. 
However, most of these methods assume positive or unsigned edges, which may impoverish the separation power \cite{yan2021two}. Recently, several studies \cite{chien2020adaptive,bo2021beyond} have applied message passing over negative (signed propagation) edges to preserve high-frequency signal exchanges. From the viewpoint of gradient flow, \cite{di2022graph} shows that negative eigenvalues can intensify high-frequency signals during propagation. On the other hand, \cite{luo2021learning} prevents message diffusion by assigning zero-weights (blocking information) to heterophilic edges to alleviate local smoothing. Based on this observation, we raise a question: \textit{does signed propagation always attain the best performance on heterophilic graphs?}

To address the above question, we focus on the recent investigations \cite{ma2021homophily,yan2021two} that provide valuable insights on message-passing. These studies mathematically analyze the effect of unsigned and signed messages on separability by comparing node feature distributions before and after message passing and aggregation. Furthermore, they reveal that signed messaging always enhances performance on heterophilic graphs.
However, our empirical analysis (Fig. \ref{status}) shows that this assertion may not hold well under real-world graphs. This discrepancy arises because the prior work assumes binary class graphs  \cite{baranwal2023effects}, impeding their extension to multi-class scenarios. In this paper, we deal with signed messaging on this condition. We first demonstrate that blind application of signed messaging to multi-class graphs incurs smoothing and performance degradation. To overcome the problem, we propose to utilize two types of calibration \cite{guo2017calibration,wang2021confident}, which are simple yet effective in enhancing the performance of signed GNNs.
In summary, our contributions can be described as follows:

\begin{itemize}
    \item Contrary to the prior work confined to binary class graphs, we address the issues when the signed messaging mechanism is extended to a multi-class scenario. Our work provides fundamental insight into using signed messages and establishes the theoretical background for developing powerful GNNs.
    \item We hypothesize and prove that signed propagation reduces the discrimination power in certain cases while increasing prediction uncertainties. Expanding this, we propose novel strategies based on two types of calibration that are shown to be effective for both signed and unsigned GNNs.
    \item We perform extensive experiments with real-world benchmark datasets to validate our theorems. Our experiments demonstrate the effectiveness of the proposed techniques.
\end{itemize}

\section{Related Work}
\textbf{Graph Neural Networks (GNNs).} Aggregating the information from adjacent nodes, GNNs have shown great potential under semi-supervised learning settings. Early study \cite{defferrard2016convolutional}
focused on the spectral graph analysis (e.g., Laplacian decomposition) in a Fourier domain. However, it suffers from large computational costs as the graph scale increases. GCN \cite{kipf2016semi} reduced the overhead by harnessing the localized spectral convolution through the first-order approximation of a Chebyshev polynomial. Another notable approach is spatial-based GNNs \cite{velickovic2017graph,brody2021attentive,morris2023orbit,barbero2023locality} which aggregate information in a Euclidean domain. Early spatial techniques have led to the development of many powerful schemes that encompass remote nodes as neighbors. 

\textbf{GNNs on Heterophilic Graphs.} Traditional message-passing GNNs fail to perform well in heterophilic graphs \cite{pei2020geom}. To redeem this problem, recent studies have focused on handling disassortative edges \cite{choi2022finding,zhao2023graph,chanpuriya2022simplified,choi2023gread,bi2024make,mao2024demystifying} by capturing node differences or by incorporating similar remote nodes as neighbors. More recently, H$_2$GCN \cite{zhu2020beyond} suggests the separation of ego and neighbors during aggregation. As another branch, selecting neighbors from non-adjacent nodes \cite{li2022finding}, configuring path-level pattern \cite{ijcai2022p310}, finding a compatibility matrix \cite{zhu2021graph}, employing adaptive propagation \cite{wang2022powerfulb}, and choosing appropriate architectures \cite{zheng2023auto} have been recently proposed. Some methodologies change the sign of disassortative edges from positive to negative  \cite{chien2020adaptive,bo2021beyond,fang2022polarized,guo2022clenshaw,lee2023towards,zheng2023auto,yan2024trainable,li2024pc} while others assign zero-weights to disassortative edges \cite{luo2021learning}. Even though \cite{baranwal2021graph} demonstrates that signed propagation can effectively handle binary class graphs, further investigation may be needed before applying the same techniques to multi-class graphs.

\section{Preliminary} \label{preliminary}
Let $\mathcal{G}=(\mathcal{V},\mathcal{E}, X)$ be a graph with $\vert\mathcal{V}\vert=n$ nodes and $\vert\mathcal{E}\vert=m$ edges. The node attribute matrix is $X \in \mathbb{R}^{n \times F}$, where $F$ is the dimension of an input vector. Given $X$, the hidden representation of node features $H^{(l)}$ at the \textit{l-th} layer is derived through message-passing. Here, node $i's$ feature is defined as $h_i^{(l)}$.
The structural property of $\mathcal{G}$ is represented by its adjacency matrix $A \in \{0, 1\}^{n \times n}$ and diagonal matrix $D$ of node degrees is derived from $A$ as $d_{ii}=\sum^n_{j=1}{A_{ij}}$.
Each node has its label $Y \in \mathbb{R}^{n \times C}$ ($C$ represents the number of classes). 
Lastly, the global edge homophily ratio, $\mathcal{H}_\mathcal{G}$, is defined as:
\begin{equation}
\label{global_homo}
\mathcal{H}_\mathcal{G} \equiv \frac{\sum_{(i,j)\in \mathcal{E}} 1(Y_i=Y_j)}  {|\mathcal{E}|}
\end{equation}
Likewise, the local homophily ratio of node $i$ ($b_i$) is given by:
\begin{equation}
\label{local_homo}
b_i \equiv \frac{{\sum^n_{j=1} A_{ij} \cdot 1(Y_i=Y_j)}} {d_{ii}}
\end{equation}
Given a partially labeled training set $\mathcal{V}_L$, the goal of semi-supervised node classification is to correctly predict the classes of unlabeled nodes $\mathcal{V}_U=\{\mathcal{V}-\mathcal{V}_L\} \subset \mathcal{V}$.

\section{Theoretical Analysis} 
We first discuss the mechanism of GNNs in terms of message-passing and parameter update (\S \ref{general_gnn}). Then, we introduce previous analyses of signed message passing on binary class graphs (\S \ref{binary_sign}) and highlight some misunderstandings revealed through an empirical study with real-world datasets (\S \ref{emp_analysis}). Based on this observation, we extend the prior analysis to a multi-class scenario and point out some drawbacks of signed messaging in general environments with multiple classes (\S \ref{problems}).

\subsection{Message-Passing Neural Networks} \label{general_gnn}
GNNs employ alternate steps of propagation and aggregation, during which the node features are updated iteratively. This is widely known as message-passing, which can be represented as follows:
\begin{equation}
\label{gnn}
\begin{gathered}
H^{(l+1)}=\phi(\bar{H}^{(l+1)})
\end{gathered}
\end{equation}
Here, $H^{(0)}=X$ is the initial vector and $H^{(l)}$ is nodes' hidden representations at the \textit{l-th} layer. $\bar{H}^{(l+1)}=AH^{(l)}W^{(l)}$ is retrieved through message-passing ($A$) and we obtain $H^{(l+1)}$ after an activation function $\phi$. $W^{(l)}$ is trainable weight matrices shared by all nodes.
The label prediction is produced by applying cross-entropy $\sigma(\cdot)$ (e.g., log-softmax) to $\bar{H}^{(L)}$, where the loss function is defined as below:
\begin{equation}
\label{loss_gnn}
\mathcal{L}_{GNN}=\mathcal{L}_{nll}(Y, \widehat{Y})
\end{equation}

The parameters are updated by computing negative log-likelihood loss $\mathcal{L}_{nll}$ between the predictions ($\widehat{Y}=\sigma(\bar{H}^{(L)})$) and true labels ($Y$).
Since most GNNs assume homophily, they tend to preserve the low-frequency information (local smoothing) \cite{nt2019revisiting}.
As a result, they fail to pinpoint the differences between heterophilic neighbors \cite{oono2019graph,pei2020geom}. To solve this, recent studies \cite{chien2020adaptive,bo2021beyond,yan2021two,chen2022does} focus on the extraction of high-frequency signals by flipping the sign of disassortative edges from positive to negative. In the following sections, we analyze the influence of information propagation from the perspectives of \textbf{message-passing} (Eq. \ref{gnn}) and \textbf{parameter update} (Eq. \ref{loss_gnn}).

\subsection{Using Signed Messages on Binary Classes} \label{binary_sign}
Before delving into the multi-class graphs, we assess the implications of three message-passing schemes (Eq. \ref{plane_mp} - \ref{prune_mp_eq}) under binary class graphs \cite{baranwal2021graph,yan2021two}.

\textit{\textbf{Message-passing} (Signed propagation always has a better discrimination power than other propagation methods)} 
For brevity, we take GCN \cite{kipf2016semi} as an example to inspect how message passing alters node feature vectors.
Assuming a binary label ($y_i \in$ \{0, 1\}), let us inherit several assumptions and notations from \cite{yan2021two} as follows: (1) For all nodes $i=\{1,...,n\}$, their degrees $\{d_i\}$ and features $\{h_i\}$ are \textit{i.i.d.} random variables. 
(2) Every class has the same population.
(3) With a slight abuse of notation, assume that $h^{(0)}=XW^{(0)}$ is the first layer projection of the initial node features. 
(4) The node feature of label $y_i$ follows the distribution ($\mu$ or $-\mu$) as,
\begin{equation}
\label{expect}
\begin{gathered}
\mathbb{E}(h^{(0)}_i|y_i) \sim
    \begin{cases}
        \mu, & \text{if  } y_i=0\\
        -\mu, & \text{if  } y_i=1
    \end{cases}
\end{gathered}
\end{equation}
We first introduce the equations from prior work \cite{yan2021two} (Eq. \ref{plane_mp} and Eq. \ref{signed_mp}), which demonstrate the updates in the expectations of an ego node $i$'s latent features $h^{(0)}_i$ after a single-hop application of message passing $h^{(1)}_i$ below:
\begin{flalign}
\label{basic}
h^{(1)}_i={h^{(0)}_i \over d_i+1} + \sum_{j \in \mathcal{N}_i}{h^{(0)}_j \over \sqrt{(d_i+1)(d_j+1)}} 
\end{flalign}

\begin{figure}[t]
     \centering
     \begin{subfigure}[b]{0.234\textwidth}
         \centering
         \includegraphics[width=\textwidth]{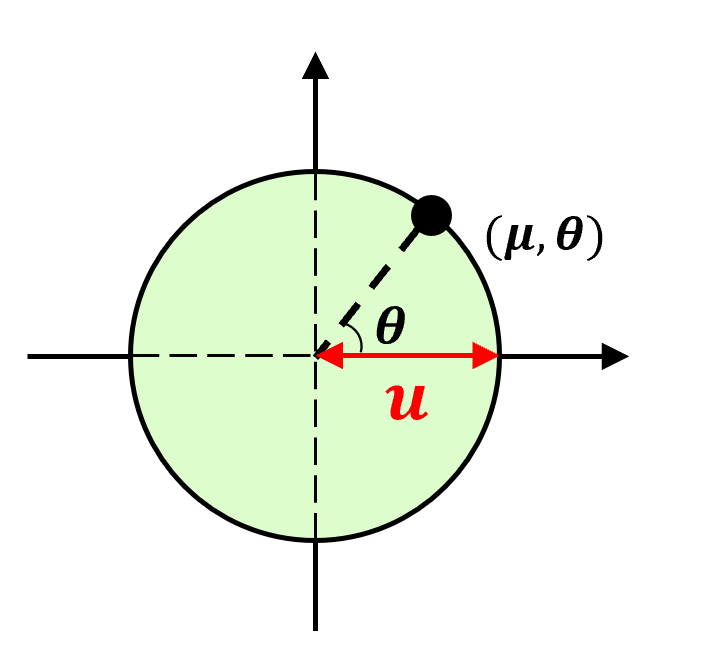}
         \caption{Binary class (C=2)}
         \label{ex_a}
     \end{subfigure}
     \begin{subfigure}[b]{0.234\textwidth}
         \centering
         \includegraphics[width=\textwidth]{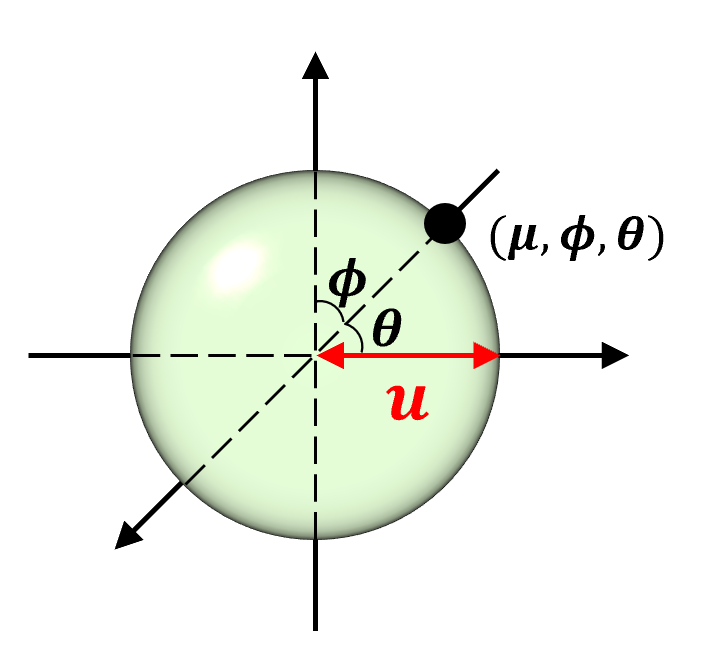}
         \caption{Multiple classes (C=3)}
         \label{ex_b}
     \end{subfigure}
        \caption{We provide an example to illustrate the distribution of node features in (a) binary and (b) multi-class scenarios. Without loss of generality, in the case of having $C$ classes, the representation can be achieved by adding $C-2$ angles to the binary class. In both cases, one can easily infer that the scale of the aggregated neighbors is always smaller than $|\mu|$, considering the Laplacian- or attention-based aggregation methods (please see Appendix B-1 for more details)}
        \label{example}
\end{figure}

\textbf{(Vanilla GCN)} As illustrated in Figure \ref{example}a (binary class), we assume $h_i \sim N(\mu,0)$ for $y_i=0$. Otherwise, $h_i \sim N(-\mu,0)$. Let us assume that $d_{ij}=\sqrt{(d_i+1)(d_j+1)}$.
Then, the updated node feature $\mathbb{E}(h^{(1)}_i|v_i,d_i)$ derived through the message-passing of vanilla GCN (Eq. \ref{basic}) is as follows:
\begin{equation}
\label{plane_mp}
\mathbb{E}(h^{(1)}_i|v_i,d_i)=\left({1+(2b_i-1)d'_i \over d_i+1}\right)\mu,   \,\,\, d'_i=\sum_{j \in \mathcal{N}_i}{1 \over d_{ij}}
\end{equation}

\textbf{(Signed GCN)}
To define the mechanism of signed GCN, we need to assume an error ratio $e$ that stands for the accuracy ($1-e$) of configuring the sign of heterophilous edges. Considering that the sign of heterophilous nodes is flipped inaccurately with a probability ($e$), the update of the signed GCN can be defined as below:
\begin{equation}
\label{signed_mp}
\mathbb{E}_s(h^{(1)}_i|v_i,d_i)= \left({1+(1-2e)d'_i \over d_i+1}\right)\mu
\end{equation}
Now, we first introduce the expectation changes of zero-weight GCN (blocking message), similar to the ones discussed above.

\textbf{(Zero-weight GCN)} Given an error ratio $e$, assigning zero weights to heterophilic edges leads to the following update in the expectation of feature distribution:
\begin{equation}
\label{prune_mp_eq}
\mathbb{E}_z(h^{(1)}_i|v_i,d_i)= \left({1+(b_i-e)d'_i \over d_i+1}\right)\mu.
\end{equation}

$\bullet$ \textit{Derivation of the above equations is in Appendix A-1} $\sim$ \textit{A-3.}

\begin{figure}
     \centering
     \begin{subfigure}[b]{0.234\textwidth}
         \centering
         \includegraphics[width=\textwidth]{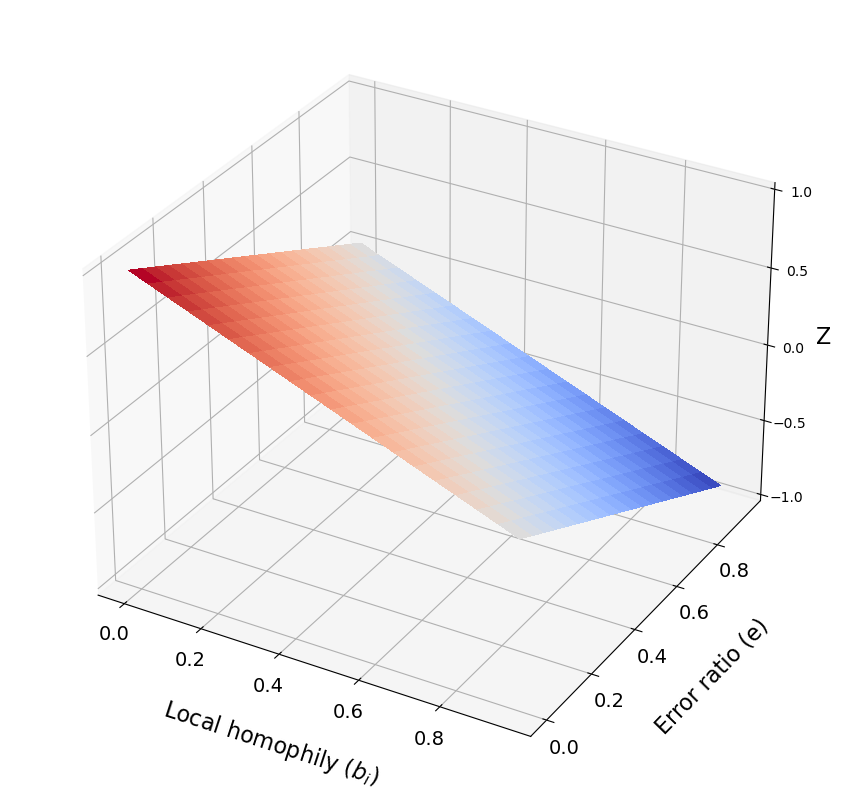}
         \caption{$Z$ in binary class}
         \label{xy_a}
     \end{subfigure}
     \begin{subfigure}[b]{0.234\textwidth}
         \centering
         \includegraphics[width=\textwidth]{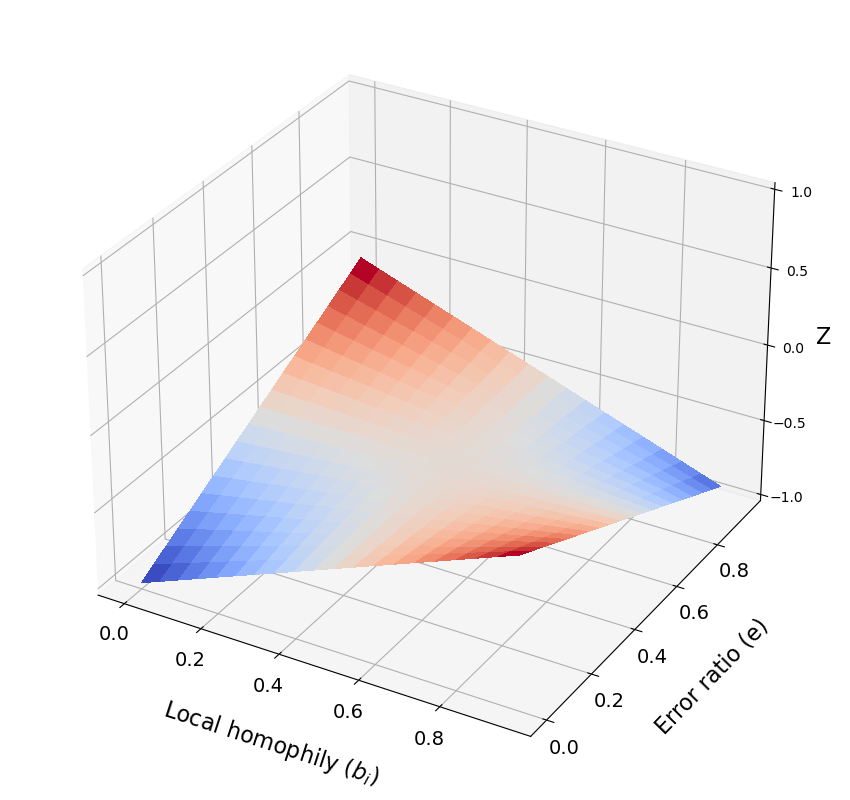}
         \caption{$Z$ in multiple classes}
         \label{xy_b}
     \end{subfigure}
        \caption{We plot the Z in (a) Eq. \ref{binary_diff} and (b) Eq. \ref{class_prob} to compare the discrimination powers of signed GCN and zero-weight GCNs. The red and blue colored parts indicate the regions where signed GCN and zero-weight GCN produce superior performances, respectively}
        \label{xy_plot}
\end{figure}

For all message-passing schemes, if the coefficient of the right equations is smaller than 1, node feature vectors move towards the decision boundary and message-passing loses its discrimination power \cite{yan2021two}.
Based on this, we compare the separability of signed GCN and zero-weight GCN below.
\begin{manualcoroll}{4.1} [Comparison of discrimination powers of signed GCN and zero-weight GCN]
\label{coroll_1}
Omitting the overlapping part of Eq. \ref{signed_mp} and Eq. \ref{prune_mp_eq}, their difference $ Z = \mathbb{E}_s(\cdot) - \mathbb{E}_z(\cdot)$ can be derived using the error ratio and the local homophily ratio as,
\begin{equation}
\label{binary_diff}
\begin{gathered}
Z=(1-2e) - (b_i-e)=1-e-b_i,
\end{gathered}
\end{equation}
where $0 \leq e, b_i \leq 1$. 
\end{manualcoroll}
In Figure \ref{xy_a}, we visualize the value $Z$ in Eq. \ref{binary_diff}. Note that the space is half-divided because $\mathbb{E}[Z]=\int_{0}^{1}\int_{0}^{1} (1-e-b_i) \,de \, db_i = 0$. 
Since $Z$ is likely to be positive when $e$ is small, it suggests that signed GCN outperforms zero-weight GCN. Likewise, we can induce the separability gap of plain (Eq. \ref{plane_mp}) and signed messaging (Eq. \ref{signed_mp}) as $Z=2(b_i+e-1)$, which is likely to be negative when the error ratio is small. Therefore, the equation indicates the superiority of signed propagation compared to other methods. In summary, we can conclude that signed messaging may be a suitable choice for binary graphs.

\textit{\textbf{Parameter update} (Signed messages consistently improve separability in binary classification scenario)} Let us assume that an ego node ($i$) and one of its neighbor nodes ($j$) are connected with a signed edge. Please ignore other neighbors and concentrate on the separation effect of signed propagation. To compute the loss function, we need to predict the label of the ego node ($i$) as follows:
\begin{equation} \label{eq_start} \begin{gathered} \widehat{Y}_i=\sigma(\bar{H}_i^{L})=\sigma\left(\frac{\bar{H}_i^{(L)}}{d_i+1} - \frac{\bar{H}_j^{(L)}}{\sqrt{(d_i+1)(d_j+1)}}\right) \end{gathered} \end{equation}
Assume that the true label of $i$ is $Y_i =k$. Then, the loss (Eq. \ref{loss_gnn}) between $Y_i$ and the prediction ($\widehat{Y}_i$) can be computed as:
\begin{equation} \label{y_hat} \mathcal{L}_{nll}(Y_i,\widehat{Y}i)=-\log(\widehat{y}{i,k}) 
\end{equation}
Accordingly, the update procedure for nodes $i$ and $j$ follow $\widehat{y}^{(t+1)}_{i,k}=\widehat{y}^t_{i,k}-\eta\nabla_i\mathcal{L}_{nll}(Y_i,\widehat{Y}_i)$ and  $\widehat{y}^{(t+1)}_{j,k}=\widehat{y}^t_{j,k}-\eta\nabla_j\mathcal{L}_{nll}(Y_i,\widehat{Y}_i)$, respectively. Here, $\eta$ is the learning rate and $\nabla$ represents a partial derivative. Then, the gradient of node $j$, $\nabla_j\mathcal{L}_{nll}(Y_i,\widehat{Y}_i)_k$ follows:
\begin{flalign}
\label{eq_end}
\nabla_j\mathcal{L}_{nll}(Y_i,\widehat{Y}_i)_k= -{1 \over \widehat{y}_{i,k}}\cdot (\widehat{y}_{i,k}(1-\widehat{y}_{i,k})\underset{\nabla_j=\partial i / \partial j}{\underline{\textbf{(-1)}}}) > 0
\end{flalign}
Since the column-wise components of the projection matrices ($W$) used to compute $\bar{H}$ (Eq. \ref{gnn}) act as independent classifiers, as the training epoch ($t$) progresses, the probability of class $k$ for node $j$ ($\widehat{y}_{j,k}$) decreases compared to node $i$ since $\nabla_i\mathcal{L}_{nll}(Y_i,\widehat{Y}_i)<0.$

$\bullet$ \textit{More details of Eq. \ref{eq_end} can be found in Appendix A-4.}

\begin{figure}[t]
 \includegraphics[width=.48\textwidth]{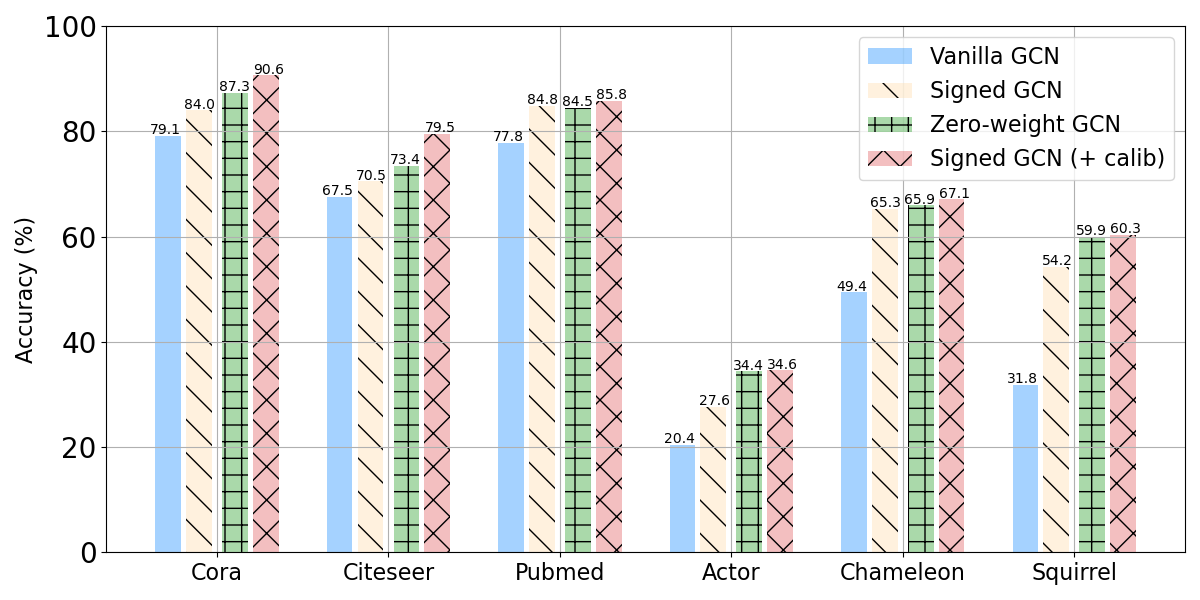}
    \caption{Node classification accuracy on six benchmark datasets. Vanilla GCN utilizes the original graph, while the coefficient of heterophilous edges is changed to -1 in signed GCN and to 0 in zero-weight GCN, respectively. Here, signed GCN (+ calib) further employs two types of calibration, which is introduced in Section \ref{methodology}}
  \label{status}
\end{figure}

\subsection{Empirical Analysis} \label{emp_analysis}
Through Corollary \ref{coroll_1} and Eq. \ref{eq_end}, we prove that signed propagation is generally advantageous when it is applied to binary class graphs.
Now, we examine whether this finding is valid on multi-class graphs with more than two classes. 
In Figure \ref{status}, we measure the node classification accuracy of GCN \cite{kipf2016semi} on nine benchmark graphs (the statistical details of the datasets are shown in Table \ref{dataset}). Starting from the original GCN called vanilla GCN, we construct two variants; one variant replaces heterophilic edges with -1 (called signed GCN), and the other assigns weight 0 to heterophilic edges (called zero-weight GCN).
Since we identified all heterophilous edges, the error ratio is zero. In this case, $Z$ $(=1-b_i)$ derived in Corollary \ref{coroll_1} has a non-negative value regardless of the local homophily ratio $b_i$, meaning that signed GCN outperforms zero-weight GCN. 
However, Figure \ref{status} shows that zero-weight GCN generally outperforms signed GCN ($Z \leq 0$).
To further investigate this phenomenon, we extend the theorems to the multi-class scenario and identify some drawbacks. We then introduce two types of calibration to enhance separability while reducing uncertainty.


\subsection{Signed Messaging on Multiple Classes} \label{problems}
Through empirical studies, we have confirmed that prior theories do not hold well in multi-class benchmark graphs. We analyze the functioning of signed message propagation on multi-class graphs and uncover several drawbacks from the perspectives of message-passing and parameter updates.

\textit{\textbf{Message-passing} (Signed propagation may cause impairments in the discrimination power of message-passing)}
Let us consider a graph with three classes. Without loss of generality, we can extend the same assumptions applied in the binary class case (Eq. \ref{expect}) to the ternary class case by employing the spherical coordinate as follows:
\begin{equation}
\label{multi_expect}
\begin{gathered}
\mathbb{E}(h^{(0)}_i|y_i) \sim (\mu, \phi,\theta)
\end{gathered}
\end{equation}
where $\mu$ represents the scale of a vector. The direction is determined by two angles $\phi$ and $\theta$. As shown in Figure \ref{example}b, this equation satisfies the origin symmetry as in the binary case since $(\mu,\pi/2,0)=(-\mu,\pi/2,0)$. Based on this, we introduce a novel understanding by extending Eq. \ref{plane_mp} and Eq. \ref{signed_mp} as follows:

\begin{manuallem}{4.2} [Signed GCN] \label{multi_flip}
Let us assume $y_i=0$. For simplicity, we assume the coordinates of the ego $(\mu,\pi/2,0)$ as $k$, and denote the aggregated neighbors as $k'$ = $(\mu',\phi', \theta')$, where $0 \leq \mu' \leq \mu$, $\phi'=\pi / 2$, and $0 \leq \theta' \leq 2\pi$ as explained in Figure \ref{example}. Then, the expectation of $h^{(1)}_i$ is defined to be: 
\begin{flalign}
\label{mu_range}
\mathbb{E}^m_s(h^{(1)}_i|v_i,d_i)= {(1-2e)\{b_ik+(b_i-1)k'\}d'_i+k \over d_i+1}
\end{flalign}
\end{manuallem}
Given the scale of all vectors is $\mu$, the aggregated vector $k'$ always satisfies $|k'| \leq \mu$ considering the degree-normalization and homophily coefficient ($1-b_i \leq 1$).

\begin{manuallem}{4.3} [Zero-weight GCN] \label{multi_obs}
Likewise, the $h^{(1)}_i$ driven by zero-weight GCN is:
\begin{flalign}
\mathbb{E}^m_z(h^{(1)}_i|v_i,d_i)={\{(1-e)b_ik+e(1-b_i)k'\}d'_i+k \over d_i+1}
\end{flalign}
\end{manuallem}

$\bullet$ \textit{Derivation can be found in Appendix B-1} $\sim$ \textit{B-2.}

Similar to the Corollary \ref{coroll_1}, we can compare the separability of the two methods under multiple classes.
\begin{manualcoroll}{4.4} [Discrimination power] \label{coroll_2}
The difference of separation powers $Z=\mathbb{E}^m_s(\cdot)-\mathbb{E}^m_z(\cdot)$ between signed GCN and zero-weight GCN in multi-class graphs is given by:
\begin{flalign}
\label{eq_disc}
Z &= (1-2e)\{b_ik+(b_i-1)k'\} - \{(1-e)b_ik+e(1-b_i)k'\} \nonumber \\
&=-eb_ik+(1-e)(b_i-1)k'
\end{flalign}
Accordingly, we can induce the following conditional statement:
\begin{equation}
\label{class_prob}
\begin{gathered}
Z \in
    \begin{cases}
        (1-e-b_i)k, & \text{if}\,\,\, k'=-k\\
        \{-2eb_i-(1-e-b_i)\}k, & \text{if}\,\,\, k'=k
    \end{cases}
\end{gathered}
\end{equation}
$\bullet$ More details are introduced in Appendix B-3.
\end{manualcoroll}

Similar to the binary class, as the distribution of aggregated neighbors $k'$ diverges from $k$ (e.g., $k' = -k$), $Z$ $(= 1 - e - b_i)$ becomes identical to Eq. \ref{binary_diff}, which converges to zero as follows:
\begin{equation}
\int_{0}^{1}\int_{0}^{1} (1-e-b_i) \,de \, db_i = 0
\end{equation}
This means selecting a proper propagation scheme depends on $b_i$ and $e$. However, as $k'$ gets closer to $k$, we can infer that $Z$ $(=-2eb_i+e+b_i-1)$ tends to be negative as below: 
\begin{equation}
\label{neg_val}
\int_{0}^{1}\int_{0}^{1} (-2eb_i+e+b_i-1) \,de \, db_i = -1
\end{equation}
where signed propagation may achieve dismal performance. Intuitively, the probability of being $\cos(k',k)=-1$ is inversely proportional to the number of entire classes. Thus, we conclude that zero-weight GCN generally outperforms signed one under multi-class datasets. As shown in Fig. \ref{status}, signed GCN outperforms zero-weight GCN only on the Pubmed dataset which has the smallest number of classes (Table \ref{status}) among the datasets. Thus, we propose to manually adjust the messages from the neighbors ($k'$) based on their similarity, denoted as edge weight calibration. The details are introduced in Section \ref{edge_calib}, which guides $k'$ to get closer to $-k$ as in the upper case of Eq. \ref{class_prob} ($Z=1-e-b_i$).

\textit{\textbf{Parameter update} (Although signed propagation contributes to the ego-neighbor separation, it also increases the prediction uncertainty under multiple class environments)}
Uncertainty management, which is closely related to the entropy \cite{liu2022confidence}, is vital in securing confident predictions \cite{moon2020confidence,mukherjee2020uncertainty}.  
Here, we focus on the conflicting evidence \cite{perry2013machine,zhao2020uncertainty} which ramps up the entropy of outputs.
One can easily measure the uncertainty of a prediction ($\widehat{y}_i$) using Shannon's entropy \cite{shannon1948mathematical} as below:
\begin{equation}
\label{entropy}
H(\widehat{y}_i)=-\sum^C_{j=1}\widehat{y}_{i,j}log_c\widehat{y}_{i,j}
\end{equation}
Now, we deduce a theorem regarding the uncertainty below. 

\begin{figure}[t]
     \centering
     \begin{subfigure}[b]{0.22\textwidth}
         \centering
         \includegraphics[width=\textwidth]{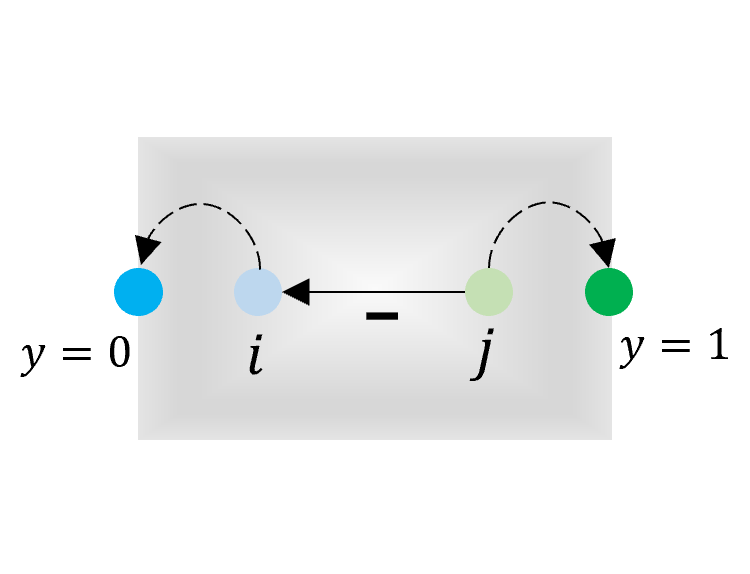}
         \caption{}
         \label{mu_a}
     \end{subfigure}
     \hfill
     \begin{subfigure}[b]{0.22\textwidth}
         \centering
         \includegraphics[width=\textwidth]{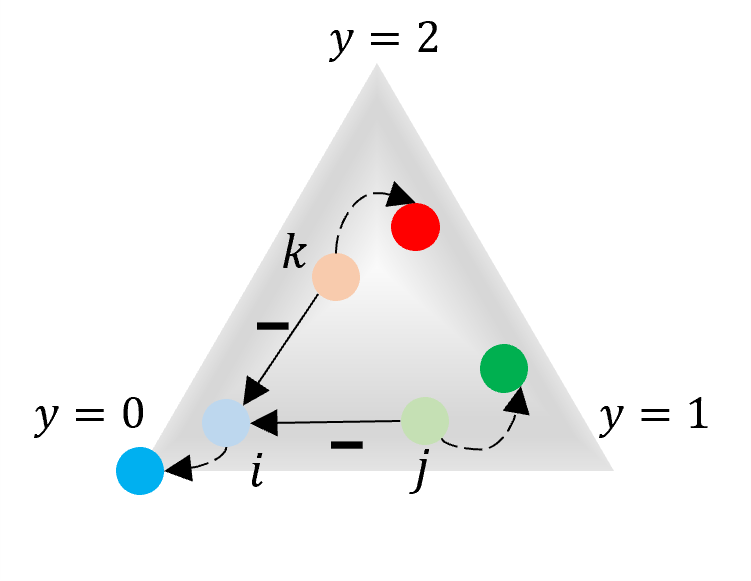}
         \caption{}
         \label{mu_b}
     \end{subfigure}
        \caption{Visualization of parameter update using the Dirichlet distribution (central side means higher prediction uncertainty). (a) Binary class case and (b) Multi-class case. In both cases, signed messages separate the ego and neighbors. However, in (b) multi-class case, the uncertainty of neighbors ($j$ and $k$) connected with signed edges increases}
        \label{mu}
\end{figure}

\begin{manualcoroll}{4.5} [Prediction Uncertainty]
\label{sign_uncertainty}
Signed messaging contributes to ego-neighbor separation (Eq. \ref{eq_end}). Nonetheless, in the case of multiple classes, the expected prediction uncertainty when using the signed propagation, $H(\mathbb{E}[\widehat{y}_s])$, remains higher than that of positive/zero-weighted, $H(\mathbb{E}[\widehat{y}_p])$, as the training epoch ($t$) progresses:
\begin{equation}
\label{last_eq}
\begin{gathered}
\lim_{t\to\infty} \{ H(\mathbb{E}[\widehat{y}^t_{s}]) - H(\mathbb{E}[\widehat{y}^t_{p}]) \} > 0
\end{gathered}
\end{equation}
$\bullet$ \textit{The proof is given in Appendix C.}
\end{manualcoroll}

For ease of understanding, in Figure \ref{mu}, we visualize the impact of signed propagation using the Dirichlet distribution. In both figures, signed messages contribute to the separation. However, in Fig \ref{mu_b}, the simultaneous increase in other class probabilities leads to the increment of conflict evidence. The solution to this will be introduced in $\S$ \ref{sec_calib} as a confidence calibration.


\section{Methodology} \label{methodology}
We introduced the limitations of signed propagation in the multi-class scenario from two perspectives; message-passing and parameter update. Here, we propose two types of calibration, both of which belong to the self-training method \cite{guo2017calibration,yang2021rethinking}, respectively. 

\subsection{(Message-Passing) Edge Weight Calibration} \label{edge_calib}
In Corollary \ref{coroll_2}, we analyze how signed propagation affects the distribution of neighbors ($k'$). Particularly, we observe that the separability degrades when $k'$ becomes similar to $k$. This observation suggests a mechanism to reduce the influence of heterogeneous neighbors. Thus, we propose the edge weight calibration based on the similarity of the two nodes, e.g., cosine similarity of $l$-th layer node features, $\cos(h^l_i,h^l_j)$. Then, we normalize the similarity score to make it belong to [0, 1], followed by multiplying it with the original edge weight $(A_{ij})$ as follows:
\begin{equation}
\forall (i,j)\in A, \,\, \tilde{A}_{ij} = 
    {\cos (h^l_i,h^l_j)+1 \over 2} \cdot A_{ij} 
\end{equation}
We should also consider the sign of a message from multi-hop away nodes, which may be inconsistent through which information is transmitted. For example, we assume three nodes $i, j, k$ that are connected serially and their edge weights are $s_{ij}$ and $s_{jk}$, respectively. If $y_i=y_j$, then $s_{ij}=1$, otherwise $s_{ij}=-1$. Under binary class, multi-hop edge weight $s_{ij} \cdot s_{jk}$ is always consistent with the node label. However, under multiple classes, if the labels of three nodes are $y_i = 0$, $y_j = 1$, and $y_k = 2$, $k$ is trained to be similar to $i$ even though $i$ and $k$ are in different classes since $s_{ij} \cdot s_{jk} = -1 \times -1 =1$. To address this, we modify the above equation using Jumping Knowledge \cite{xu2018representation,lei2022evennet}, assuming $l$-hop message-passing as shown below:
\begin{equation}
\label{method_one}
\forall (i,j)\in \sum^L_{l=1} A^L, \,\, A_{ij} = 
    {\cos (h^l_i,h^l_j)+1 \over 2} \cdot A_{ij} 
\end{equation}
Note that any type of $A_{ij}$ (e.g., attention values) can be used for calibration. Below, we prove that edge weight calibration can enhance the separability of GNNs.

\begin{manualtheorem}{5.1} [Discrimination power after edge weight calibration] \label{thm5.1} Assuming that every class has the same population, the expectations of aggregated neighbors in signed GNN with and without the edge weight calibration are denoted as $\mathbb{E}\left[|k'_{edge}|\right]$ and $\mathbb{E}\left[|k'|\right]$, respectively. Then, the two expectations comply with the following inequality relation. 
\begin{equation}
\label{eq4.9}
\begin{gathered}
\mathbb{E}\left[|k'_{edge}|\right]={|k| \over d}\sqrt{{c \over 4}+ \sum^c_{j=0}\cos({2\pi j \over c})^2} > \mathbb{E}\left[|k'|\right]
\end{gathered}
\end{equation}
where we prove that an increase in distance from the decision boundary $\mathbb{E}\left[|k'_{edge}|\right]>\mathbb{E}\left[|k'|\right]$ indicates improved separability.
\end{manualtheorem}
$\bullet$ \textit{Proof is provided in Appendix D.}

Below, we demonstrate that edge weight calibration also improves the performance of \textit{positive GNNs}, such as GCN \cite{kipf2016semi} and GAT \cite{velickovic2017graph}, by reflecting the cosine similarity of two nodes \cite{brody2021attentive}.

\subsubsection{Edge weight calibration for positive GNN}
We explain how edge weight calibration improves the quality of positive GNNs. For brevity, we take the aggregation scheme of GCN \cite{kipf2016semi}. Then, the expectation of node $i$ after message-passing $h^{(1)}_i$ becomes,
\begin{equation}
\label{unsigned_gnn}
\mathbb{E}(h^{(1)}_i|v_i,d_i)= {k \over d_i+1} + {\{kb_i + k'(1-b_i)\}d'\over d_i+1} 
\end{equation}
Since the edge weight calibration multiplies the normalized cosine similarity $0 \leq s \leq 1$, the expectation of calibrated GCN can be redefined as follows:
\begin{flalign}
\label{calib_gcn}
\mathbb{E}(h^{(1)}_i|v_i,d_i)={k \over d_i+1} + {\{(s-1)ke+k\}b_i + (1-s)k'e(1-b_i)\over d_i+1}d'
\end{flalign}
Using the above two equations, we can derive the discrimination power between the calibrated (Eq. \ref{calib_gcn}) and original GCN (Eq. \ref{plane_mp}) as,
\begin{flalign}
    Z = (1-s)keb_i+\{1+(s-1)e\}k'(1-b_i) 
\end{flalign}
Then, the conditional statement is given by:
\begin{equation}
\begin{gathered}
Z \in
    \begin{cases}
        e+b_i-1-se & \text{if}\,\, \cos(k',k)=-1\\
        e+b_i-1-se+2eb_i(s-1) & \text{if}\,\, \cos(k',k)=1,
    \end{cases}
\end{gathered}
\label{pos_sep}
\end{equation}
Since $\int_{0}^{1}\int_{0}^{1} (e+b_i-1) \,de db = \left[1-{e^2 + b^2 \over 2}\right]^1_{e,b=0} = 0$, we can induce that the first line of the above equation is smaller than 0. In addition, if we assume that $s$ is sufficiently small ($s \rightarrow 0$), the second line of Eq. \ref{pos_sep} simplifies to $e+b_i-1-2eb_i$ (Eq. \ref{neg_val}), indicating that edge weight calibration improves the separability of positive GNNs.

$\bullet$ \textit{Proof and calibrated GAT \cite{velickovic2017graph} can be found in Appendix E-1.}

\begin{table*}
\caption{(Q1) Node classification accuracy (\%) with standard deviation on the six test datasets. \textbf{Bold} and indicates the $1^{st}$ and gray cells stand for the top-3 performances. Values in brackets represent the dissonance (Eq. \ref{dissonance}) and methods with calibrations are marked with $\ddagger$. The average improvement with calibration is 2.8\% compared to the original method} 
\label{perf_1}
\centering
\begin{center}
\begin{adjustbox}{width=.97\textwidth}
\begin{tabular}{@{}llllllll}
\Xhline{2\arrayrulewidth}
        & Datasets                       & Cora & Citeseer & Pubmed & Actor & Chameleon & Squirrel \\ 
        & $\mathcal{H}_g$ (Eq. \ref{global_homo}) & 0.81 & 0.74 & 0.8 & 0.22 & 0.23 & 0.22 \\
\Xhline{2\arrayrulewidth}
                        & GCN \cite{kipf2016semi}              & 79.0 $_{\,\pm\,0.6 }$ \footnotesize{(0.17)}  & 67.5 $_{\,\pm\,0.8 }$ \footnotesize{(0.29)} & 77.6 $_{\,\pm\,0.2 }$ \footnotesize{(0.53)} & 20.2 $_{\,\pm\,0.4 }$  \footnotesize{(0.29)} & 49.3 $_{\,\pm\,0.5 }$ \footnotesize{(0.19)} & 30.7 $_{\,\pm\,0.7 }$ \footnotesize{(0.31)}  \\
                        & \textbf{GCN}$^\ddagger$       & 81.0 $_{\,\pm\,0.9 }$ \footnotesize{(0.12)} & 71.3 $_{\,\pm\,1.2 }$ \footnotesize{(0.14)} & 77.8 $_{\,\pm\,0.4 }$ \footnotesize{(0.38)} & 21.7 $_{\,\pm\,0.6 }$ \footnotesize{(0.62)} & 49.4 $_{\,\pm\,0.6 }$ \footnotesize{(0.25)} & 31.5 $_{\,\pm\,0.6 }$ \footnotesize{(0.58)} \\
                        & GAT \cite{velickovic2017graph}            & 80.1 $_{\,\pm\,0.6 }$ \footnotesize{(0.22)} & 68.0 $_{\,\pm\,0.7 }$ \footnotesize{(0.25)} & 78.0 $_{\,\pm\,0.4 }$ \footnotesize{(0.45)} & 22.5 $_{\,\pm\,0.3 }$ \footnotesize{(0.28)} & 47.9 $_{\,\pm\,0.8 }$ \footnotesize{(0.17)} & 30.8 $_{\,\pm\,0.9 }$ \footnotesize{(0.27)} \\
                        & \textbf{GAT}$^\ddagger$             & 81.4 $_{\,\pm\,0.4 }$ \footnotesize{(0.12)} & 72.2 $_{\,\pm\,0.6 }$ \footnotesize{(0.08)} & 78.3 $_{\,\pm\,0.3 }$ \footnotesize{(0.39)} & 23.2 $_{\,\pm\,1.8 }$ \footnotesize{(0.43)} & 49.2 $_{\,\pm\,0.4 }$ \footnotesize{(0.16)} & 30.3 $_{\,\pm\,0.8 }$ \footnotesize{(0.40)} \\
                        & GIN \cite{xu2018powerful}             & 77.3 $_{\,\pm\,0.8 }$ \footnotesize{(0.33)} & 66.1 $_{\,\pm\,0.6 }$ \footnotesize{(0.29)} & 77.1 $_{\,\pm\,0.7 }$ \footnotesize{(0.47)} & 24.6 $_{\,\pm\,0.8 }$ \footnotesize{(0.51)} & 49.1 $_{\,\pm\,0.7 }$ \footnotesize{(0.26)} & 28.4 $_{\,\pm\,2.2 }$ \footnotesize{(0.48)} \\
                        & APPNP \cite{klicpera2018predict}                & 81.3 $_{\,\pm\,0.5 }$ \footnotesize{(0.15)} & 68.9 $_{\,\pm\,0.3 }$ \footnotesize{(0.21)} & 79.0 $_{\,\pm\,0.3 }$ \footnotesize{(0.42)} & 23.8 $_{\,\pm\,0.3 }$ \footnotesize{(0.49)} & 48.0 $_{\,\pm\,0.7 }$ \footnotesize{(0.34)} & 30.4 $_{\,\pm\,0.6 }$ \footnotesize{(0.69)} \\
                        & GCNII \cite{chen2020simple}          & 81.1 $_{\,\pm\,0.7 }$ \footnotesize{(0.08)} & 68.5 $_{\,\pm\,1.4 }$ \footnotesize{(0.13)} & 78.5 $_{\,\pm\,0.4 }$ \footnotesize{(0.20)} & 25.9 $_{\,\pm\,1.2 }$ \footnotesize{(0.43)} & 48.1 $_{\,\pm\,0.7 }$ \footnotesize{(0.21)} & 29.1 $_{\,\pm\,0.9 }$ \footnotesize{(0.24)} \\
                        & H$_2$GCN  \cite{zhu2020beyond}      & 80.6 $_{\,\pm\,0.6 }$ \footnotesize{(0.16)} & 68.2 $_{\,\pm\,0.7 }$ \footnotesize{(0.22)} & 78.5 $_{\,\pm\,0.3 }$ \footnotesize{(0.29)} & 25.6 $_{\,\pm\,1.0 }$ \footnotesize{(0.34)} & 47.3 $_{\,\pm\,0.8 }$ \footnotesize{(0.19)} & 31.3 $_{\,\pm\,0.7 }$ \footnotesize{(0.62)} \\
                        & ACM-GCN \cite{luan2022revisiting} & 80.2 $_{\,\pm\,0.8 }$ \footnotesize{(0.24)} & 68.3 $_{\,\pm\,1.1 }$ \footnotesize{(0.17)} & 78.1 $_{\,\pm\,0.5 }$ \footnotesize{(0.31)} & 24.9 $_{\,\pm\,2.0 }$ \footnotesize{(0.46)} & 49.5 $_{\,\pm\,0.7 }$ \footnotesize{(0.20)} & 31.6 $_{\,\pm\,0.4 }$ \footnotesize{(0.54)} \\
                        & HOG-GCN \cite{wang2022powerfulb} & 79.7 $_{\,\pm\,0.4 }$ \footnotesize{(0.31)} & 68.2 $_{\,\pm\,0.6 }$ \footnotesize{(0.24)} & 78.0 $_{\,\pm\,0.2 }$ \footnotesize{(0.29)} & 21.5 $_{\,\pm\,0.5 }$ \footnotesize{(0.37)} & 47.7 $_{\,\pm\,0.5 }$ \footnotesize{(0.32)} & 30.1 $_{\,\pm\,0.4 }$ \footnotesize{(0.51)}  \\
                        & JacobiConv \cite{wang2022powerfula} & 81.9 $_{\,\pm\,0.6 }$ \footnotesize{(0.26)} & 69.6 $_{\,\pm\,0.8 }$ \footnotesize{(0.19)} & 78.5 $_{\,\pm\,0.4 }$ \footnotesize{(0.24)} & 25.7 $_{\,\pm\,1.2 }$ \footnotesize{(0.30)} & \cellcolor{Gray}\textbf{52.8 $_{\,\pm\,0.9 }$} \footnotesize{(0.21)} & 32.0 $_{\,\pm\,0.6 }$ \footnotesize{(0.37)} \\
                        & GloGNN \cite{li2022finding} & 82.4 $_{\,\pm\,0.3 }$ \footnotesize{(0.33)} & 70.3 $_{\,\pm\,0.5 }$ \footnotesize{(0.35)} & \cellcolor{Gray}79.3 $_{\,\pm\,0.2 }$ \footnotesize{(0.26)} & \cellcolor{Gray}26.6 $_{\,\pm\,0.7 }$ \footnotesize{(0.38)} & 48.2 $_{\,\pm\,0.3 }$ \footnotesize{(0.27)} & 28.8 $_{\,\pm\,0.8 }$ \footnotesize{(0.43)} \\
                        & AERO-GNN \cite{lee2023towards} & 81.6 $_{\,\pm\,0.5 }$ \footnotesize{(0.25)} & 71.1 $_{\,\pm\,0.6 }$ \footnotesize{(0.44)} & 79.1 $_{\,\pm\,0.5 }$ \footnotesize{(0.20)} & 25.5 $_{\,\pm\,1.1 }$ \footnotesize{(0.51)} & 49.8 $_{\,\pm\,2.3 }$ \footnotesize{(0.41)} & 29.9 $_{\,\pm\,1.9 }$ \footnotesize{(0.66)} \\
                        & Auto-HeG \cite{zheng2023auto} & 81.5 $_{\,\pm\,1.1 }$ \footnotesize{(0.19)} & 70.9 $_{\,\pm\,1.4 }$ \footnotesize{(0.20)} & 79.2 $_{\,\pm\,0.2 }$ \footnotesize{(0.21)} & 26.1 $_{\,\pm\,1.0 }$ \footnotesize{(0.27)} & 48.7 $_{\,\pm\,1.4 }$ \footnotesize{(0.26)} & 31.5 $_{\,\pm\,1.1 }$ \footnotesize{(0.38)} \\
                        & TED-GCN \cite{yan2024trainable} & 81.8 $_{\,\pm\,0.9 }$ \footnotesize{(0.37)} & 71.4 $_{\,\pm\,0.6 }$ \footnotesize{(0.26)} & 78.6 $_{\,\pm\,0.3 }$ \footnotesize{(0.44)} & 26.0 $_{\,\pm\,0.9 }$ \footnotesize{(0.41)} & 50.4 $_{\,\pm\,1.2 }$ \footnotesize{(0.28)} & \cellcolor{Gray}\textbf{33.0} $_{\,\pm\,0.9 }$ \footnotesize{(0.50)}\\
                        & PCNet \cite{li2024pc} & 81.5 $_{\,\pm\,0.8 }$ \footnotesize{(0.28)} & 71.2 $_{\,\pm\,1.2 }$ \footnotesize{(0.31)} & 78.8 $_{\,\pm\,0.3 }$ \footnotesize{(0.33)} & 26.4 $_{\,\pm\,0.8 }$ \footnotesize{(0.27)} & 48.1 $_{\,\pm\,1.7 }$ \footnotesize{(0.30)} & 31.4 $_{\,\pm\,0.6 }$ \footnotesize{(0.36)} \\
\Xhline{2\arrayrulewidth}
                        & PTDNet \cite{luo2021learning}         & 81.2 $_{\,\pm\,0.9 }$ \footnotesize{(0.24)} & 69.5 $_{\,\pm\,1.2 }$ \footnotesize{(0.42)} & 78.8 $_{\,\pm\,0.5 }$ \footnotesize{(0.44)} & 21.5 $_{\,\pm\,0.6 }$ \footnotesize{(0.33)} & 50.6 $_{\,\pm\,0.9 }$ \footnotesize{(0.17)} & \cellcolor{Gray}32.1 $_{\,\pm\,0.7 }$ \footnotesize{(0.34)}\\
                        & \textbf{PTDNet}$^\ddagger$      & 81.9 $_{\,\pm\,0.6 }$ \footnotesize{(0.20)} & 71.1 $_{\,\pm\,0.8 }$ \footnotesize{(0.31)} & 79.0 $_{\,\pm\,0.2 }$ \footnotesize{(0.38)} & 22.7 $_{\,\pm\,0.6 }$ \footnotesize{(0.19)} & \cellcolor{Gray}50.9 $_{\,\pm\,0.3 }$ \footnotesize{(0.15)} & \cellcolor{Gray}32.3 $_{\,\pm\,0.5 }$ \footnotesize{(0.30)} \\
\Xhline{2\arrayrulewidth}
                        & GPRGNN \cite{chien2020adaptive} & 82.2 $_{\,\pm\,0.4 }$ \footnotesize{(0.25)} & 70.4 $_{\,\pm\,0.8 }$ \footnotesize{(0.43)} & 79.1 $_{\,\pm\,0.1 }$ \footnotesize{(0.26)} & 25.4 $_{\,\pm\,0.5 }$ \footnotesize{(0.55)} & 49.1 $_{\,\pm\,0.7 }$ \footnotesize{(0.25)} & 30.5 $_{\,\pm\,0.6 }$ \footnotesize{(0.36)} \\
                        & \textbf{GPRGNN}$^\ddagger$ & \cellcolor{Gray}\textbf{84.7 $_{\,\pm\,0.2 }$} \footnotesize{(0.04)} & \cellcolor{Gray}73.3 $_{\,\pm\,0.5 }$ \footnotesize{(0.05)} & \cellcolor{Gray}\textbf{80.2} $_{\,\pm\,0.2 }$ \footnotesize{(0.11)} & \cellcolor{Gray}\textbf{28.1} $_{\,\pm\,1.3 }$ \footnotesize{(0.33)} & \cellcolor{Gray}51.0 $_{\,\pm\,0.4 }$ \footnotesize{(0.18)} & 31.8 $_{\,\pm\,0.4 }$ \footnotesize{(0.16)} \\
                        & FAGCN \cite{bo2021beyond}       & 81.9 $_{\,\pm\,0.5 }$ \footnotesize{(0.15)} & 70.8 $_{\,\pm\,0.6 }$ \footnotesize{(0.17)} & 79.0 $_{\,\pm\,0.5 }$ \footnotesize{(0.31)} & 25.2 $_{\,\pm\,0.8 }$ \footnotesize{(0.66)} & 46.5 $_{\,\pm\,1.1 }$ \footnotesize{(0.25)} & 30.4 $_{\,\pm\,0.4 }$ \footnotesize{(0.64)} \\
                        & \textbf{FAGCN}$^\ddagger$ & \cellcolor{Gray}84.1 $_{\,\pm\,0.4 }$ \footnotesize{(0.09)} & \cellcolor{Gray}\textbf{73.8 $_{\,\pm\,0.5 }$} \footnotesize{(0.08)} & \cellcolor{Gray}79.7 $_{\,\pm\,0.2 }$ \footnotesize{(0.16)} & \cellcolor{Gray}27.6 $_{\,\pm\,0.5 }$ \footnotesize{(0.42)} & 48.8 $_{\,\pm\,0.7 }$ \footnotesize{(0.13)} & 31.3 $_{\,\pm\,0.5 }$ \footnotesize{(0.37)} \\
                        & GGCN \cite{yan2021two} & 81.0 $_{\,\pm\,1.2 }$ \footnotesize{(0.38)} & 70.7 $_{\,\pm\,1.6 }$ \footnotesize{(0.30)} & 78.2 $_{\,\pm\,0.4 }$ \footnotesize{(0.47)} & 22.5 $_{\,\pm\,0.5 }$ \footnotesize{(0.47)} & 48.5 $_{\,\pm\,0.7 }$ \footnotesize{(0.15)} & 30.2 $_{\,\pm\,0.7 }$ \footnotesize{(0.40)} \\
                        & \textbf{GGCN}$^\ddagger$ & \cellcolor{Gray}83.9 $_{\,\pm\,0.8 }$ \footnotesize{(0.07)} & \cellcolor{Gray}73.0 $_{\,\pm\,0.4 }$ \footnotesize{(0.05)} & 78.9 $_{\,\pm\,0.3 }$ \footnotesize{(0.29)} & 24.6 $_{\,\pm\,0.4 }$ \footnotesize{(0.26)} & 50.0 $_{\,\pm\,0.4 }$ \footnotesize{(0.07)} & 31.1 $_{\,\pm\,0.6 }$ \footnotesize{(0.15)} \\
\Xhline{2\arrayrulewidth}
\end{tabular}
\end{adjustbox}
\end{center}
\end{table*}

\begin{table}[t]
\caption{Statistical details of the six benchmark datasets}
\label{dataset}
\centering
\begin{adjustbox}{width=0.48\textwidth}
\begin{tabular}{@{}llllllll}
&     &        &         &  & & & \\ 
\Xhline{2\arrayrulewidth}
        & Datasets         & Cora  & Citeseer & Pubmed & Actor & Cham. & Squirrel \\ 
\Xhline{2\arrayrulewidth}
                        & \# Nodes  & 2,708  & 3,327   & 19,717 & 7,600 & 2,277  & 5,201 \\
                        & \# Edges         & 10,558  & 9,104  & 88,648   & 25,944 & 33,824  & 211,872 \\
                        & \# Features       & 1,433  & 3,703  & 500   & 931 & 2,325  & 2,089 \\
                        & \# Labels        & 7  & 6  & 3     & 5  & 5  & 5 \\
\Xhline{2\arrayrulewidth}
\end{tabular}
\end{adjustbox}
\end{table}

\subsection{(Parameter Update) Confidence Calibration} \label{sec_calib}
In Corollary \ref{sign_uncertainty}, we proved that signed messages increase the uncertainty of predictions. To redeem this, we propose the calibration of predictions. This scheme boasts several advantages such as being free from path configuration, cost-efficient, and powerful. To begin with, we claim that various loss functions can be used for confidence calibration as below:
\begin{equation}
\label{calib_loss}
\begin{gathered}
\mathcal{L}_{conf} \in
    \begin{cases} 
    {1 \over N}\sum_{i=1}^N(-\text{max}(\widehat{y}_i)+\text{submax}(\widehat{y}_i)) \\
    \sum_{i=1}^N ||I-\widehat{y}_i \widehat{y}_i^T||
    \end{cases}
\end{gathered}
\end{equation}
The first approach suppresses the maximum values and reduces the differences between the maximum and sub-maximum values, while the second is a well-known orthogonal constraint. Both can be used to alleviate conflicting evidence in predictions. It is important to note that any penalization strategy that reduces entropy can be applied as a method of confidence calibration. Some may argue that this is similar to prior schemes \cite{wang2021confident,guo2022orthogonal}. However, we offer a new perspective: signed propagation may introduce higher uncertainty (Thm. \ref{sign_uncertainty}), and we observe a significant performance improvement without using the labels of validation nodes. The theorem below demonstrates that confidence calibration can significantly reduce the uncertainty of signed GNNs.
\begin{manualtheorem}{5.2} [Reduced uncertainty]
\label{reduced_uncertainty}
Given the specific training step ($t$) and global homophily ($\mathcal{H}_g$ in Eq. \ref{global_homo}), entropy of signed propagation with confidence calibration $H(\mathbb{E}[\widehat{y}^t_{cal}])$ (upper side of Eq. \ref{calib_loss}) is smaller than using signed messages only $H(\mathbb{E}[\widehat{y}^t_{s}])$. This can be induced by extending the Corollary \ref{sign_uncertainty} as follows:
\begin{equation}
\label{reduced_eq}
\begin{gathered}
\lim_{t\to\infty} H(\mathbb{E}[\widehat{y}^t_{cal}]) \approx \mathcal{H}_g \cdot \underset{\text{calibration}}{\underline{\log ({c-1 \over 1+\lambda})}} < \mathcal{H}_g \cdot \log (c-1) = H(\mathbb{E}[\widehat{y}^t_s])
\end{gathered}
\end{equation}
\end{manualtheorem}

$\bullet$ \textit{The proof is provided in Appendix E-2} ($\lambda > 0$ is $lr$ in Eq. \ref{total_loss}).

\subsection{Optimization}
During the training phase, we apply the edge (Eq. \ref{method_one}) and confidence calibration (Eq. \ref{calib_loss}) to reduce the smoothing and uncertainty as, 
\begin{equation}
\label{total_loss}
\mathcal{L}_{total}=\mathcal{L}_{EGNN}+\lambda\mathcal{L}_{conf}
\end{equation}
Here, $\mathcal{L}_{EGNN}$ represents the node classification loss with the edge weight calibration, and $\lambda$ is a hyper-parameter that adjusts the intensity of the confidence calibration. As in Figure \ref{status}, the enhanced performance of signed GCN (with calibration) substantiates the significant performance improvement attributable to our methodology. Additionally, note that this strategy is agnostic to specific GNN methodologies, which enables broad applicability across other frameworks \cite{chien2020adaptive,bo2021beyond,yan2021two} since it leverages only node similarity and predictions. In Section \ref{lambda_analysis}, we conduct experiments to select the $\lambda$ that achieves the best validation score for each dataset. 

$\bullet$ \textit{Please refer to the pseudo-code and time complexity in Appendix F. Also, our code is in anonymous GitHub\footnote{https://anonymous.4open.science/r/Signed-Calibrated-GNN-34B6/README.md} for reproducibility}.

\section{Experiments} \label{experiments}
We conduct extensive experiments to address the following research questions. We measure the node classification accuracy of all methods following the settings in \cite{kipf2016semi}, where the parameters are optimized solely with the training nodes.
\begin{itemize}
    \item \textbf{Q1} Do the proposed methods improve the node classification accuracy of graph neural networks?
    \item \textbf{Q2} Do the signed messages cause an increase in the uncertainty of the final prediction?
    \item \textbf{Q3} How much impact do the two calibration methodologies have on performance improvement?
    \item \textbf{Q4} How does the weight of the confidence calibration, denoted as $\lambda$ (Eq. \ref{total_loss}) affect the overall performance?
\end{itemize}

\textbf{Baselines:} We employ various algorithms with (1) positive edge weights: GCN \cite{kipf2016semi}, GAT \cite{velickovic2017graph}, GIN \cite{xu2018powerful}, APPNP \cite{klicpera2018predict}, GCNII \cite{chen2020simple}, H$_2$GCN \cite{zhu2020beyond}, ACM-GCN \cite{luan2022revisiting}, HOG-GCN \cite{wang2022powerfulb}, JacobiConv \cite{wang2022powerfula}, GloGNN \cite{li2022finding}, TED-GCN \cite{yan2024trainable} and PCNet \cite{li2024pc}, (2) zero-weighted: PTDNet \cite{luo2021learning}, and (3) three methods with signed propagation: GPRGNN \cite{chien2020adaptive}, FAGCN \cite{bo2021beyond}, and GGCN \cite{yan2021two}.

$\bullet$ \textit{Please refer to Appendix G for datasets and implementation.}

\subsection{Experimental Results (Q1)} \label{ex_result}
In Table \ref{perf_1}, we describe the node classification accuracy of each method. Please note that all performances in Table \ref{perf_1} can be inferior to the benchmarks since we chose the test scores with the best validation.
We also show dissonance within parentheses. Dissonance is an uncertainty metric, a measure of effectiveness in distinguishing out-of-distribution data from conflict predictions \cite{zhao2020uncertainty,huang2022end}:
\begin{equation}
\label{dissonance}
diss(\widehat{y}_i)=\sum^C_{j=1}\left(\widehat{y}_{ij}\sum_{k\neq j}\widehat{y}_{ik}(1-{|\widehat{y}_{ik}-\widehat{y}_{ij}|\over \widehat{y}_{ij}+\widehat{y}_{ik}})\over \sum_{k\neq j}\widehat{y}_{ik}\right)
\end{equation}
The above equation can be computed for the non-zero values of $\widehat{y}_i$. Now, we analyze the results from the two perspectives below.

(1) Relationship between homophily ratio and performance: On three homophilic citation networks (Cora, Citeseer, and Pubmed), we observe that all methods including plain, zero-weighted, and signed GNNs generally perform well. In addition, several algorithms such as JacobiConv, GloGNN, and AERO-GNN, achieve impressive results under these conditions. As the homophily ratio decreases, methods that adjust weights outperform positive message-passing approaches. For instance, GloGNN and JacobiConv excel compared to simpler models on the Actor dataset. Similarly, several signed GNNs (GPRGNN, FAGCN, and GGCN) and ego-neighbor separation method (H$_2$GCN) demonstrate strong performance in our experiments. Notably, on two heterophilic graphs (Chameleon and Squirrel), which contain many cyclic edges, blocking information propagation (PTDNet) outperforms other GNNs.

(2) Calibrations are effective in improving accuracy as well as in alleviating uncertainty: We apply calibrations ($\ddagger$) to six methods that employ positive (GCN, GAT), zero-weights (PTDNet), and signed propagation (GPRGNN, FAGCN, and GGCN). The average improvements achieved by applying calibrations to these three methods are 2.8\%, demonstrating that calibrations significantly enhance the accuracy of generic GNNs. Specifically, GPRGNN$^\ddagger$ shows remarkable improvements, becoming the top performer across multiple datasets, with accuracy increases to 84.7\% on Cora, 73.3\% on Citeseer, and 80.2\% on Pubmed. As hypothesized throughout this paper, calibration is effective for signed messaging. For example, GPRGNN$^\ddagger$ and FAGCN$^\ddagger$ outperform their uncalibrated versions on highly heterophilic datasets such as Chameleon and Squirrel, emphasizing the impact of calibration in multi-class settings. Additionally, we observe that the calibrated methods show lower dissonance (Eq. \ref{dissonance}) compared to the corresponding vanilla models. This indicates that our method not only refines the smoothing effect but also reduces uncertainty by mitigating conflicting signals.

$\bullet$ \textit{In Appendix H-1, we present a case study to evaluate the impact of the number of classes on dissonance (uncertainty). Additionally, experiments on node classification accuracy with more baselines and datasets are provided in Appendix H-2.}

\begin{figure}[t]
 \includegraphics[width=.48\textwidth]{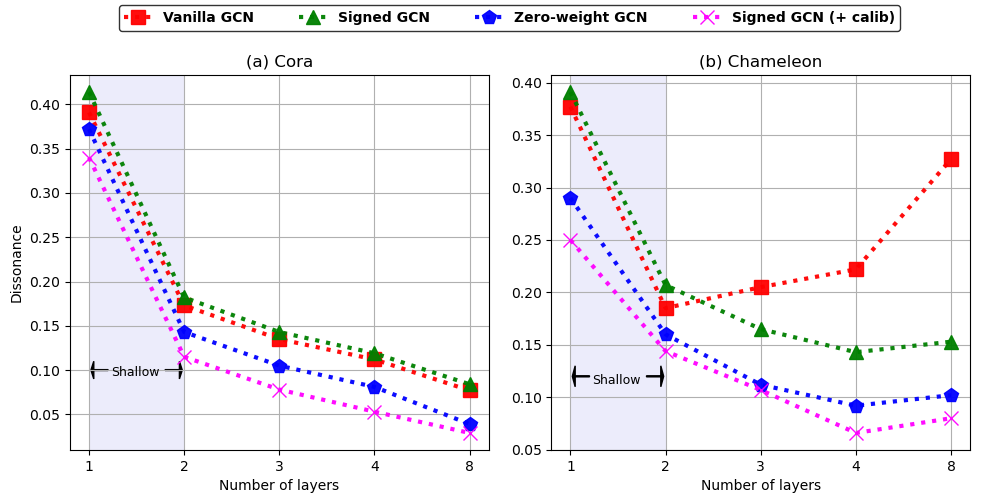}
    \caption{(Q2) Dissonance of vanilla GCN and its two variants; signed and zero-weight, which is the same as the one in Figure \ref{status}. The left one is \textit{Cora} and the right one is the \textit{Chameleon}}
  \label{val_dist}
\end{figure}

\subsection{Signed Propagation and Uncertainty (Q2)}
To demonstrate the effect of signed messaging on uncertainty, we measure the dissonance using two variants of GCN; one variant assigns -1 and the other assigns zeros to heterophilic edges, respectively (please refer to \S \ref{emp_analysis} for details). The results are presented in Fig. \ref{val_dist} where the x-axis is the number of layers and the y-axis represents dissonance. Here, we conducted experiments with the Cora (Fig. 4a) and Chameleon (Fig. 4b) datasets.
As stated in Thm. \ref{sign_uncertainty}, the signed GCN variant in the green-colored line exhibits higher uncertainty compared to the zero-weight variant. In the Chameleon dataset, the entropy of vanilla GCN increases as the layer gets deeper, indicating that positive message-passing fails to deal adequately with heterophily. Although zero-weight GCN shows lower dissonance, signed GCN with calibrations consistently produces the smallest uncertainty, highlighting the efficacy of our method in mitigating such limitations.

\begin{table}[h]
\caption{(Q3) Node classification accuracy with standard deviation by applying either the edge weight calibration or the confidence calibration to signed GNNs}
\label{q3}
\centering
\begin{adjustbox}{width=0.48\textwidth}
\begin{tabular}{@{}llllll}
& \multicolumn{1}{l}{} &     &        &         &  \\ 
\Xhline{2\arrayrulewidth}
        & Datasets         & Cora  & Citeseer & Actor & Chameleon\\ 
\Xhline{2\arrayrulewidth}
                        & GPRGNN  & 82.2 $_{\,\pm\,0.4 }$  & 70.4 $_{\,\pm\,0.8 }$  & 25.4 $_{\,\pm\,0.5 }$ & 49.1 $_{\,\pm\,0.9 }$ \\
                        & \textbf{{\footnotesize\,\,+ edge calib}}     & 83.3 $_{\,\pm\,0.6 }$  & 71.5 $_{\,\pm\,1.0 }$  & 26.3 $_{\,\pm\,0.6 }$ & 49.7 $_{\,\pm\,0.7 }$  \\
                        & \textbf{{\footnotesize\,\,+ conf calib}}       & \textbf{84.1} $_{\,\pm\,0.5 }$ & 72.6 $_{\,\pm\,0.5 }$  & \textbf{27.5} $_{\,\pm\,0.4 }$ & \textbf{50.2} $_{\,\pm\,0.3 }$   \\
\Xhline{2\arrayrulewidth}
                        & FAGCN        & 81.9 $_{\,\pm\,0.5 }$  & 70.8 $_{\,\pm\,0.6 }$  & 25.2 $_{\,\pm\,0.8 }$ & 46.5 $_{\,\pm\,1.1 }$ \\
                        & \textbf{{\footnotesize\,\,+ edge calib}}       & 82.8 $_{\,\pm\,0.6 }$ & 72.3 $_{\,\pm\,0.7 }$ & 25.8 $_{\,\pm\,0.7 }$ & 46.9 $_{\,\pm\,1.0 }$ \\
                        & \textbf{{\footnotesize\,\,+ conf calib}}        & 83.5 $_{\,\pm\,0.3 }$ & \textbf{73.4} $_{\,\pm\,0.5 }$  & 26.3 $_{\,\pm\,0.4 }$ & 48.0 $_{\,\pm\,0.8 }$ \\
\Xhline{2\arrayrulewidth}
                        & GGCN        & 81.0 $_{\,\pm\,1.2 }$ & 70.7 $_{\,\pm\,1.6 }$ & 22.5 $_{\,\pm\,0.5 }$  & 48.5 $_{\,\pm\,0.7 }$  \\
                        & \textbf{{\footnotesize\,\,+ edge calib}}       & 82.9 $_{\,\pm\,1.0 }$ & 71.0 $_{\,\pm\,1.1 }$ & 23.3 $_{\,\pm\,0.6 }$ & 48.9 $_{\,\pm\,0.7 }$ \\
                        & \textbf{{\footnotesize\,\,+ conf calib}}        & 83.7 $_{\,\pm\,0.9 }$  & 72.5 $_{\,\pm\,0.6 }$  & 23.6 $_{\,\pm\,0.4 }$ & 49.6 $_{\,\pm\,0.5 }$ \\
\Xhline{2\arrayrulewidth}
\end{tabular}
\end{adjustbox}
\end{table}

\subsection{Ablation Study (Q3)} \label{sec_ablation}
We conduct an ablation study to test the relative effectiveness of the two calibration methods. In Table \ref{q3}, we describe the node classification accuracy of signed GNNs on the homophilic (Cora, Citeseer) and heterophilic (Actor, Chameleon) graphs. Here, we apply either the edge weight calibration (+ edge calib) or the confidence calibration (+ conf calib). From the table, methods with the confidence calibration generally outperform the edge weight calibration and exhibit smaller variances. This is because the confidence calibration reduces the uncertainty of all nodes during training while the edge calibration is only applied to a (small) subset of edges during testing. Also, edge calibration is beneficial, preventing the propagation of signed information from similar nodes.

\begin{figure}[t]
 \includegraphics[width=.48\textwidth]{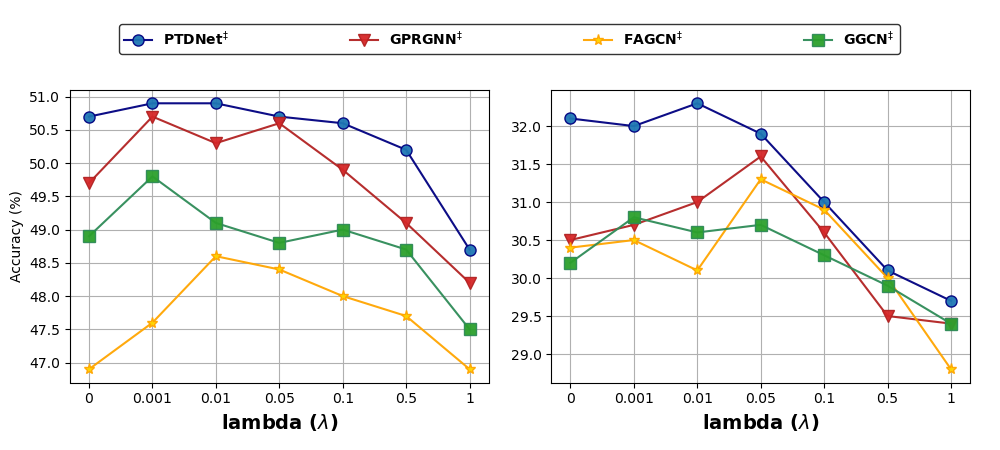}
    \caption{(Q4) Node classification accuracy w.r.t. the parameter $\lambda$ which controls the weight of the confidence calibration. Each figure represents a Chameleon (left) and Squirrel (right)}
  \label{ablation}
\end{figure}

\subsection{Parameter Sensitivity Analysis (Q4)} \label{lambda_analysis}
We conduct experiments to investigate the effect of the confidence calibration by adjusting $\lambda$ (Eq. \ref{total_loss}). In Figure \ref{ablation}, we varied the values of $\lambda$ from 0 to 1. The edge weight calibration is combined with all methods along with the confidence calibration. The blue line represents PTDNet, while the others are signed GNNs. Notably, in both figures, calibrations improve the quality of all methods and are shown to be more effective in signed GNNs than in PTDNet, particularly in terms of relieving uncertainty. Also, finding an appropriate lambda is beneficial for overall improvement since assigning higher values to $\lambda$ generally degrades the overall performance, necessitating early stopping during training. Additionally, please note that calibration can be limited by the inherent low capability of base models in heterophilic graphs.

\section{Conclusion}
We present new theoretical insights on the effect of using signed messages in multi-class graphs. In contrast to previous theorems that assume graphs with binary classes and solely focus on message-passing, we extend them to scenarios with multiple classes and introduce a new perspective on parameter updates. From this viewpoint, we highlight two critical limitations of using signed propagation: (1) it suffers from reduced separability in multi-class graphs with high possibility, and (2) it increases the probability of generating conflicting evidence compared to positive message-passing schemes. Based on these observations, we propose the edge weight and confidence calibrations as solutions to enhance performance and alleviate uncertainty. Our extensive performance experiments on six real-world datasets show that the calibration techniques are effective for signed and plain GNNs. We believe that our theorems offer valuable insights for future studies on developing improved aggregation schemes to reduce the smoothing effect.


\bibliographystyle{ACM-Reference-Format}
\bibliography{references.bib}

\onecolumn

\section*{Technical Appendix}

\begin{appendices}
$\\\\$
\textbf{A. Proof of Equation 6, 8, 9, and 13}
$\\$

\textbf{A-1. Proof of Equation 6 [Binary class, vanilla GCN]} 

More detailed proofs of Equation 6 and 8 are provided in \cite{yan2021two}. Thus, we introduce them briefly here. 
Assume a binary class $y_i \in \{0,1\}$. Using the aggregation scheme of GCN \cite{kipf2016semi}, the hidden representation of node $i$ after message-passing $h^{(1)}_i$ is defined as:
\begin{flalign}
\label{basic}
h^{(1)}_i={h^{(0)}_i \over d_i+1} + \sum_{j \in \mathcal{N}_i}{h^{(0)}_j \over \sqrt{(d_i+1)(d_j+1)}} 
\end{flalign}
As illustrated in Figure \ref{example}a (binary class), we assume $h_i \sim N(\mu,{1 \over \sqrt{d_i}})$ if $y_i=0$ and otherwise $h_i \sim N(-\mu,{1 \over \sqrt{d_i}})$. Based on the local homophily ratio $b_i$, Eq. \ref{basic} can be extended as:
\begin{flalign}
\mathbb{E}(h^{(1)}_i|v_i,d_i)&={\mu \over d_i+1} + \sum_{j \in \mathcal{N}_i}\left({b_i \over \sqrt{(d_i+1)(d_j+1)}}\mu-{(1-b_i) \over \sqrt{(d_i+1)(d_j+1)}}\mu\right) \,\,\, \\
&= \left({1 \over d_i+1} + \sum_{j \in \mathcal{N}_i}{2b_i-1 \over \sqrt{(d_i+1)(d_j+1)}}\right)\mu \\
&= \left({1 \over d_i+1} + {2b_i-1  \over d_i+1}\sum_{j \in \mathcal{N}_i}{\sqrt{d_i+1} \over \sqrt{d_j+1}}\right)\mu \\
&= \left({1+(2b_i-1)d'_i \over d_i+1}\right)\mu.
\end{flalign}

\textbf{A-2. Proof of Equation 8 [Binary class, signed GCN]}

Similarly, signed GCN correctly configures the sign of heterophilous edges with the following error ratio $1-e$. For example, the sign of heterophilous nodes changes from $-\mu$ to $\mu$ with a probability $1-e$ and vice versa:
\begin{flalign}
\mathbb{E}(h^{(1)}_i|v_i,d_i)&={\mu \over d_i+1} + \sum_{j \in \mathcal{N}_i}\left({\mu(1-e) - \mu e \over \sqrt{(d_i+1)(d_j+1)}}b_i+{\mu(1-e)-\mu e \over \sqrt{(d_i+1)(d_j+1)}}(1-b_i)\right) \\
&= {\mu \over d_i+1} + \sum_{j \in \mathcal{N}_i}\left({1-2e \over \sqrt{(d_i+1)(d_j+1)}}\mu\right) \\
&= \left({1 \over d_i+1} + {1-2e  \over d_i+1}\sum_{j \in \mathcal{N}_i}{\sqrt{d_i+1} \over \sqrt{d_j+1}}\right)\mu \\
&= \left({1+(1-2e)d'_i \over d_i+1}\right)\mu.
\end{flalign}

\textbf{A-3. Proof of Equation 9 [Binary class, zero-weight GCN]}

Likewise, we can induce the expectation of zero-weight GCN by assigning zero weights on heterophilous nodes as follows:
\begin{flalign}
\mathbb{E}(h^{(1)}_i|v_i,d_i)&={\mu \over d_i+1} + \sum_{j \in \mathcal{N}_i}\left({\mu(1-e) - \mu e \times 0 \over \sqrt{(d_i+1)(d_j+1)}}b_i+{\mu(1-e)\times 0-\mu e  \over \sqrt{(d_i+1)(d_j+1)}}(1-b_i)\right) \\
&= {\mu \over d_i+1} + \sum_{j \in \mathcal{N}_i}\left({b_i-e \over \sqrt{(d_i+1)(d_j+1)}}\mu\right) \\
&= \left({1 \over d_i+1} + {b_i-e  \over d_i+1}\sum_{j \in \mathcal{N}_i}{\sqrt{d_i+1} \over \sqrt{d_j+1}}\right)\mu \\
&= \left({1+(b_i-e)d'_i \over d_i+1}\right)\mu.
\end{flalign}

$\\\\\\$
\textbf{A-4. Proof of Equation 13}

We first show that signed messages can contribute to separating the ego from its neighbors. Let us assume the label of the ego node $i$ is $k$. A neighbor node $j$ is connected to $i$ with a signed edge. Since the column-wise components of the weight matrix act as an independent classifier, the probabilities that the two nodes belong to the same class $k$, $\widehat{y}_{i,k},\widehat{y}_{j,k}$, at a training epoch $t$ are derived as,
\begin{equation}
\label{eq_1}
\begin{gathered}
\widehat{y}^{(t+1)}_{i,k}=\widehat{y}^t_{i,k}-\eta\nabla_i\mathcal{L}_{nll}(Y_i,\widehat{Y}_i)_k \\
\widehat{y}^{(t+1)}_{j,k}=\widehat{y}^t_{j,k}-\eta\nabla_j\mathcal{L}_{nll}(Y_i,\widehat{Y}_i)_k
\end{gathered}
\end{equation}
The loss function is defined as $\mathcal{L}_{nll}(Y_i,\widehat{Y}_i)_k=-\log(\widehat{y}_{i,k}),$ \,\,\, where $\,\,\, \widehat{Y}_i=\sigma(\bar{H}_i^{L})=\sigma({\bar{H}_i^{(L)} \over d_i+1} - {\bar{H}_j^{(L)} \over \sqrt{(d_i+1)(d_j+1)}})$.
The gradient $\nabla_i\mathcal{L}_{nll}(Y_i,\widehat{Y}_i)_k$ is well-known to be,
\begin{flalign}
\label{eq3}
\nabla_i\mathcal{L}_{nll}(Y_i,\widehat{Y}_i)_k 
&= {\partial \mathcal{L}_{nll}(Y_i,\widehat{Y}_i)_k \over \partial \widehat{y}_{i,k}} 
= {\partial \mathcal{L}_{nll}(Y_i,\widehat{Y}_i)_k \over \partial \widehat{y}_{i,k}} \cdot {\partial \widehat{y}_{i,k} \over \partial h^{(L)}_{i,k}} \\
&= -{1 \over \widehat{y}_{i,k}}\cdot (\widehat{y}_{i,k}(1-\widehat{y}_{i,k}))
= \widehat{y}_{i,k}-1 < 0
\end{flalign}.
Similarly, the gradient $\nabla_j\mathcal{L}_{nll}(Y_i,\widehat{Y}_i)_k$ is given by:
\begin{flalign}
\nabla_j\mathcal{L}_{nll}(Y_i,\widehat{Y}_i)_k 
&= {\partial \mathcal{L}_{nll}(Y_i,\widehat{Y}_i)_k \over \partial \widehat{y}_{i,k}} 
= {\partial \mathcal{L}_{nll}(Y_i,\widehat{Y}_i)_k \over \partial \widehat{y}_{i,k}} \cdot {\partial \widehat{y}_{i,k} \over \partial h^{(L)}_{\textbf{j},k}} \\
&= -{1 \over \widehat{y}_{i,k}}\cdot (\widehat{y}_{i,k}(1-\widehat{y}_{i,k})(-1))
= 1-\widehat{y}_{i,k} > 0,
\end{flalign}
where we can infer that $\widehat{y}^{(t+1)}_{i,k} > \widehat{y}^{t}_{i,k}$ and $\widehat{y}^{(t+1)}_{j,k} < \widehat{y}^{t}_{j,k}$.

$\\\\$
\textbf{B. Proof of Lemma 4.2, 4.3, and Corollary 4.4}
$\\$

\textbf{B-1. Proof of Lemma 4.3 [Multi-class, signed GCN]} \label{sec_b1}

As shown in Figure \ref{example}b, we extend a binary classification scenario to a multi-class case. Without loss of generality, we employ spherical coordinates and ensure that $\mu$ corresponds to the scale of a vector, while the direction of each vector lies between zero and $\phi, \theta$ concerning their features.
Here, we assume the label is $y_i=0$. For simplicity, we replace $(\mu,\phi=\pi/2,\theta=0)$ as $k$ and $(\mu,\phi',\theta')$ as $k'$, respectively.
Though $k'$ comprises multiple distributions that are proportional to the number of classes, their aggregation always satisfies $|k'_{aggr}| \leq \mu$ since the summation of coefficients ($1-b_i$) is lower than 1 and $|k'| \leq \mu$. Referring to Fig. \ref{example}c, we can see that ${k_1+k_2 \over 2}\leq \mu$ given $b_1=b_2=0.5$, where the aggregation of neighbors always lies in $\mu$.
Thus, for brevity, we indicate $k'_{aggr}$ as $k'$ here. Now, we can retrieve the expectation ($h_i$) as follows: 
\begin{flalign}
\label{thm4.5}
\mathbb{E}(h^{(1)}_i|v_i,d_i)&={k \over d_i+1} + \sum_{j \in \mathcal{N}_i}\left({k(1-e) - k e \over \sqrt{(d_i+1)(d_j+1)}}b_i+{-k'(1-e)+k' e \over \sqrt{(d_i+1)(d_j+1)}}(1-b_i)\right) \\
&= {k \over d_i+1} + \sum_{j \in \mathcal{N}_i}\left({k(1-2e)b_i - k'(1-2e)(1-b_i)\over \sqrt{(d_i+1)(d_j+1)}}\right) \\
&= {k \over d_i+1} + \sum_{j \in \mathcal{N}_i}\left({(1-2e)\{kb_i + k'(b_i-1)\}\over \sqrt{(d_i+1)(d_j+1)}}\right) \\
&= {k \over d_i+1} + {(1-2e)\{kb_i + k'(b_i-1)\}d'\over d_i+1} \\
&= {(1-2e)\{b_ik+(b_i-1)k'\}d'_i+k \over d_i+1}.
\end{flalign}

\textbf{B-2. Proof of Lemma 4.4 [Multi-class, zero-weight GCN]}

Similar to Eq. \ref{thm4.5}, the expectation of assigning zero weights for multi-class GCN is given by:
\begin{flalign}
\label{thm4.6}
\mathbb{E}(h^{(1)}_i|v_i,d_i)&={k \over d_i+1} + \sum_{j \in \mathcal{N}_i}\left({k(1-e)  - k e \times 0 \over \sqrt{(d_i+1)(d_j+1)}}b_i+{-k'(1-e)\times 0+k' e  \over \sqrt{(d_i+1)(d_j+1)}}(1-b_i)\right) \\
&= {k \over d_i+1} + \sum_{j \in \mathcal{N}_i}\left({k(1-e)b_i + k'e(1-b_i)\over \sqrt{(d_i+1)(d_j+1)}}\right) \\
&= {k \over d_i+1} + {\{(1-e)b_ik+e(1-b_i)k'\}d'_i\over d_i+1} \\
&= {\{(1-e)b_ik+e(1-b_i)k'\}d'_i+k \over d_i+1}.
\end{flalign}

\textbf{B-3. Analysis on Corollary 4.5 [Discrimination power]}

Taking Eq. \ref{thm4.5} and \ref{thm4.6}, one can compare the discrimination power between a sign-flip and zero-weight GCN under a multi-class scenario. Exclude the overlapping part ${k \over d_i+1}$, $Z$ can be retrieved as below:
\begin{flalign}
Z &= (1-2e)\{b_ik+(b_i-1)k'\} - \{(1-e)b_ik+e(1-b_i)k'\}\\
&=-eb_ik+(1-e)(b_i-1)k'
\end{flalign}
As mentioned above, $k'$ = $(\mu',\theta')$ where $0 \leq \mu' \leq \mu$ and $0 \leq \theta' \leq 2\pi$. 
If the scale of the aggregated vector is $|\mu'|=0$, $Z=-eb_ik \leq 0$ and this implies zero-weight GCN generally outperforms signed GCN.
Instead, assuming the scale as $\mu'=\mu$, $Z$ is determined w.r.t. the angle of $k'$ as follows:
\begin{equation}
\begin{gathered}
Z \in
    \begin{cases}
        1-e-b, & \text{if}\,\, \cos(k',k)=-1\\
        -2eb-(1-e-b), & \text{if}\,\, \cos(k',k)=1,
    \end{cases}
\end{gathered}
\end{equation}
Firstly, if $k'$ is origin symmetric to $k$, the plane $Z$ is half-divided as,
\begin{equation}
\int_{0}^{1}\int_{0}^{1} (1-e-b) \,de db = \left[1-{e^2 + b^2 \over 2}\right]^1_{e,b=0} = 0 
\end{equation}
However, as $k'$ gets closer to $k$, we notice that $Z$ tends to be negative as below:
\begin{equation}
\int_{0}^{1}\int_{0}^{1} (-2eb+e+b-1) \,de db  = \left[{-eb^2 -e^2b + e^2+b^2\over 2}-1\right]^1_{e,b=0}=-1.
\end{equation}
Intuitively, the probability of being $\cos(k',k)=-1$ is inversely proportional to the number of entire classes. Thus, one can infer that zero-weight GCN generally outperforms signed one given multi-class datasets.

\textbf{C. Proof of Theorem 4.6 [Uncertainty]}

\begin{figure}[t]
    \includegraphics[width=\textwidth]{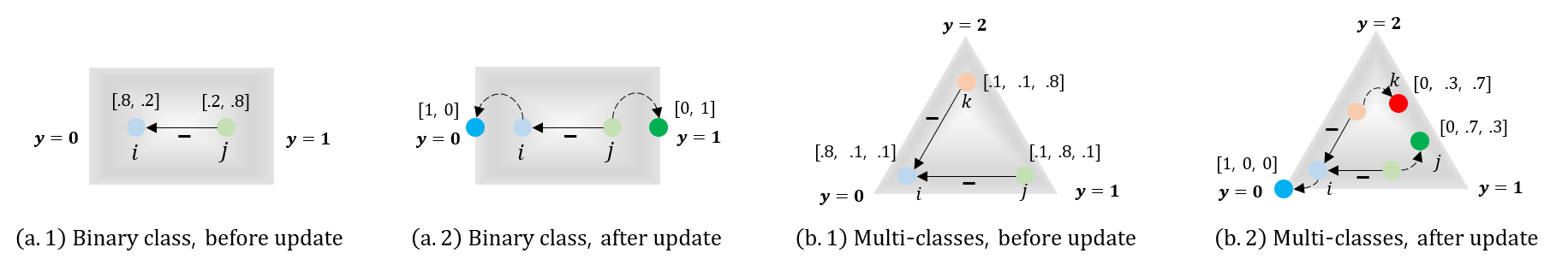}
    \caption{We visualize the update procedure of node features for (a) binary and (b) multi-class cases using Dirichlet distribution}
  \label{ex2}
\end{figure}

In Section \textbf{A.1.}, we proved that signed messages contribute to the separation of ego and neighbors as below:
\begin{flalign}
\widehat{y}^{(t+1)}_{i,k}&=\widehat{y}^t_{i,k}-\eta\nabla_i\mathcal{L}_{nll}(Y_i,\widehat{Y}_i)_k > \widehat{y}^t_{i,k}\\
\widehat{y}^{(t+1)}_{j,k}&=\widehat{y}^t_{j,k}-\eta\nabla_j\mathcal{L}_{nll}(Y_i,\widehat{Y}_i)_k < \widehat{y}^t_{j,k},
\end{flalign}

in case the label of ego ($i$) is $k$, and the neighboring node ($j$) is connected with a signed edge. $\\$

Here, let us assume two nodes $s$ and $p$, which are connected to a central node $i$ with a signed ($s$) and plain edge ($p$), respectively. We aim to show that a difference between the $H(\widehat{y}_s)$ and $H(\widehat{y}_p)$ increases w.r.t. the training epoch ($t$) under a \underline{multi-class} scenario:
\begin{equation}
\lim_{t\to\infty} \{ H(\mathbb{E}[\widehat{y}^t_{s}]) - H(\mathbb{E}[\widehat{y}^t_{p}]) \} > 0
\end{equation}

Firstly, the true label probability ($k$) of node $p$ $\widehat{y}_{p,k}$ increases, while other probabilities $\widehat{y}_{p,o}$ ($o \neq k$) decrease as follows:
\begin{equation}
\begin{gathered}
\widehat{y}^{(t+1)}_p \in
    \begin{cases} \widehat{y}^t_{p,k}-\eta\nabla_p\mathcal{L}_{nll}(Y_i,\widehat{Y}_i)_k > \widehat{y}^t_{p,k}, \\
    \widehat{y}^t_{p,o}-\eta\nabla_p\mathcal{L}_{nll}(Y_i,\widehat{Y}_i)_o < \widehat{y}^t_{p,o}, \,\, \forall \,\, o\neq k.
    \end{cases}
\end{gathered}
\end{equation}

Since we proved that $\nabla_p\mathcal{L}_{nll}(Y_i,\widehat{Y}_i)_k < 0$ in Eq. \ref{eq3}, we analyze the partial derivative $\nabla_p\mathcal{L}_{nll}(Y_i,\widehat{Y}_i)_o$ \, ($\forall$ $o \neq k$):
\begin{flalign}
\nabla_p\mathcal{L}_{nll}(Y_i,\widehat{Y}_i)_o 
&= {\partial \mathcal{L}_{nll}(Y_i,\widehat{Y}_i)_o \over \partial \widehat{y}_{p,o}} 
= {\partial \mathcal{L}_{nll}(Y_i,\widehat{Y}_i)_o \over \partial \widehat{y}_{i,o}} \cdot {\partial \widehat{y}_{i,o} \over \partial h^{(L)}_{p,o}} \\
&= {1 \over \widehat{y}_{i,o}}\cdot (\widehat{y}_{i,o}(1-\widehat{y}_{i,o}))
= 1-\widehat{y}_{i,o} > 0,
\end{flalign} 

which satisfies $\widehat{y}^t_{p,o}-\eta\nabla_p\mathcal{L}_{nll}(Y_i,\widehat{Y}_i)_o < \widehat{y}^t_{p,o}$.

On the contrary, the gradient of node $s$ has a different sign with node $p$, where we can infer that:
\begin{equation}
\begin{gathered}
\widehat{y}^{(t+1)}_s \in
    \begin{cases} \widehat{y}^t_{s,k}-\eta\nabla_s\mathcal{L}_{nll}(Y_i,\widehat{Y}_i)_k < \widehat{y}^t_{s,k},  \\
    \widehat{y}^t_{s,o}-\eta\nabla_s\mathcal{L}_{nll}(Y_i,\widehat{Y}_i)_o > \widehat{y}^t_{s,o}, \,\, \forall \,\, o\neq k.
    \end{cases}
\end{gathered}
\end{equation}

As the training epoch increases, $\widehat{y}_{p,k}$ will converge to 1 resulting in the decrease of $H(\mathbb{E}[\widehat{y}_p])$. Conversely, $\widehat{y}_{s,k}$ gets closer to 0, which may fail to generate a highly confident prediction and leads to a surge of uncertainty. Thus, one can infer that $H(\mathbb{E}[\widehat{y}^{(t+1)}_s])-H(\mathbb{E}[\widehat{y}^{(t+1)}_p]) > H(\mathbb{E}[\widehat{y}^t_s])-H(\mathbb{E}[\widehat{y}^t_p])$ as $t \rightarrow \infty$. As shown in Figure \ref{ex2}a, this can be effective under a binary class, while the signed nodes ($i,j$) in a multi-class case (Fig. \ref{ex2}b) have conflicting evidence except for class 0.
Taking another example, let us assume that the original probability (before the update) is $\widehat{y}^t_i=[0.6,0.2,0.2]$ with $C=3$. Then, one can calculate the Shannon's entropy ($E$) of $\widehat{y}^t_i$ as,
\begin{equation}
H(\widehat{y}^t_i) = -\sum^3_{j=1}\widehat{y}^t_{i,j}log_3\widehat{y}^t_{i,j} \approx 0.8649
\end{equation}

Without considering node degree, let us assume the gradient of class $k$ as $\nabla_p\mathcal{L}_{nll}(Y_i,\widehat{Y}_i)_k=-\nabla_s\mathcal{L}_{nll}(Y_i,\widehat{Y}_i)_k=\alpha$, and other classes as $\nabla_p\mathcal{L}_{nll}(Y_i,\widehat{Y}_i)_o=-\nabla_s\mathcal{L}_{nll}(Y_i,\widehat{Y}_i)_o={\alpha \over C-1}$ ($\forall$ $o \neq k$). If we take $\alpha=0.1$, $\widehat{y}^{(t+1)}_p$ and $\widehat{y}^{(t+1)}_s$ becomes:
\begin{equation}
\label{eq_74}
H(\widehat{y}^{(t+1)}_p) = H(\left[0.8,0.1,0.1\right]) \approx 0.5817, \,\,
H(\widehat{y}^{(t+1)}_s) = H(\left[0.4,0.3,0.3\right]) \approx 0.9911,
\end{equation}
where we can see that $H(\mathbb{E}[\widehat{y}^{(t+1)}_p]) < H(\mathbb{E}[\widehat{y}^{(t+1)}_s])$ as $t$ increases.

\textbf{D. Proof of Theorem 5.1 [Discrimination power after edge weight calibration]}
$\\$

We aim to show that edge weight calibration improves the signed propagation. Please recall the separability between the signed and zero-weight GNN is as below.

\begin{equation} \label{eq_d}
    Z=-eb_ik+(1-e)(b_i-1)k'.
\end{equation}

Here, $k'$ stands for a single vector that can be retrieved through the neighbor aggregation. Similarly, let us define $k'_{edge}$ as the aggregation of neighbors with the edge weight calibration. Also, referring Figure \ref{example}c, we assume that $|k'| \leq \mu$ and $|k'_{edge}| \leq \mu$ (please refer to the $\S$ B.1. for details). Since the discrimination power is determined based on the distance from the decision boundary, we can substitute $k'$ and $k'_{edge}$ in terms of $k$. Specifically, Eq. \ref{eq_d} can be redefined as below:
\begin{equation} 
\begin{gathered}
    Z=-eb_ik+(1-e)(b_i-1)\underline{\cos(k',k)|k'|}, \\
    Z_{edge}=-eb_ik+(1-e)(b_i-1)\underline{\cos(k'_{edge},k)|k'_{edge}|}.
\end{gathered}
\end{equation} \label{dist_k'}
Now, let's assume the number of classes is $c$, the average degree of nodes is $d$, and every class has the same population. Based on the above equation, one can compare the scale and direction of $\cos(k',k)|k|$ and $\cos(k'_{edge},k)|k'_{edge}|$ since other variables are the same. As $k$ is a fixed vector, the expectation of $\cos(k',k)|k'|$ can be represented as below:
\begin{equation}
\mathbb{E}\left[\cos(k',k)|k'|\right]={|k| \over d}\sqrt{\Big\{\sum^c_{j=0}\cos({2\pi j \over c})\Big\}^2+\Big\{\sum^c_{j=0}\sin({2\pi j \over c})\Big\}^2}={|k| \over d}.
\end{equation}
Since the class distribution is identical, we can imagine an n-sided polygon inscribed in a circle, and the sum of the elements of the neighbor vectors $k'$ is 0. Before retrieving the expectation of $\cos(k'_{edge},k)|k'_{edge}|$, please remember that our edge weight calibration normalized cosine similarity with the original edge weight as below:
\begin{equation}
\forall (i,j)\in \mathcal{E}, \,\, a_{ij} = {\cos (h^l_i,h^l_j)+1 \over 2} \cdot a_{ij} 
\end{equation}
Using the above equation, we can induce $\cos(k'_{edge},k)|k'_{edge}|$ as follows:
\begin{flalign}
\mathbb{E}\left[\cos(k'_{edge},k)|k'_{edge}|\right]&={|k| \over d}\sqrt{\Big\{\sum^c_{j=0}-\cos({2\pi j \over c})\cos({2\pi j \over c})\Big\}^2+\Big\{\sum^c_{j=0}-\cos({2\pi j \over c})\sin({2\pi j \over c})\Big\}^2}   \\
&= {|k| \over d}\sqrt{\Big\{\sum^c_{j=0}{-1 - \cos({4\pi j \over c}) \over 2}\Big\}^2+\Big\{\sum^c_{j=0}-{\sin({4\pi j \over c}) \over 2}\Big\}^2} \,\,\,\,\,\,\,\,\,\, (angle \,\, addition)\\
&= {|k| \over d}\sqrt{{c \over 4}+ \sum^c_{j=0}\cos({2\pi j \over c})^2}  > \mathbb{E}\left[\cos(k',k)|k'|\right]=0,
\end{flalign}
Thus, assuming that $\mathbb{E}\left[\cos(k'_{edge},k)|k'_{edge}|\right]=\alpha$ is a positive value, we can infer the following inequality:
\begin{equation}
\mathbb{E}\left[Z_{edge}\right]=\mathbb{E}\left[Z\right] + \alpha > \mathbb{E}\left[Z\right] \,\,\,\,\,\,\,\,\,\,\,\, (\alpha > 0).
\end{equation}
Even though each class has a different number of nodes, the class weight (scalar) is applied for both $k'$ and $k'_{edge}$ identically, which leads to the same conclusion. We can easily see that edge weight calibration separates $k'_{edge}$ from $k$, which completes the proof.

$\\$
\textbf{E. Edge Weight and Confidence Calibration}
$\\$

\textbf{E-1. Edge weight calibration for GCN \cite{kipf2016semi}.}
We explain how edge weight calibration improves the quality of positive GNNs. For brevity, we take the aggregation scheme of GCN \cite{kipf2016semi} as before. Then, for positive GNNs, the expectation of node $i$ after message-passing $h^{(1)}_i$ becomes:
\begin{flalign}
\mathbb{E}(h^{(1)}_i|v_i,d_i)&={k \over d_i+1} + \sum_{j \in \mathcal{N}_i}\left({k(1-e) + k e \over \sqrt{(d_i+1)(d_j+1)}}b_i+{k'(1-e)+k' e \over \sqrt{(d_i+1)(d_j+1)}}(1-b_i)\right) \\
&= {k \over d_i+1} + \sum_{j \in \mathcal{N}_i}\left({kb_i + k'(1-b_i)\over \sqrt{(d_i+1)(d_j+1)}}\right) \\
&= {k \over d_i+1} + {\{kb_i + k'(1-b_i)\}d'\over d_i+1} 
\end{flalign}
Since the edge weight calibration multiplies the normalized cosine similarity $0 \leq s \leq 1$, the expectation can be redefined as:
\begin{flalign}
\mathbb{E}(h^{(1)}_i|v_i,d_i)&={k \over d_i+1} + \sum_{j \in \mathcal{N}_i}\left({k(1-e) + k e \times s \over \sqrt{(d_i+1)(d_j+1)}}b_i+{k'(1-e)\times s+k' e  \over \sqrt{(d_i+1)(d_j+1)}}(1-b_i)\right) \\
&= {k \over d_i+1} + \sum_{j \in \mathcal{N}_i}\left({\{(s-1)ke+k\}b_i + (1-s)k'e(1-b_i)\over \sqrt{(d_i+1)(d_j+1)}}\right) \\
&= {k \over d_i+1} + {\{(s-1)ke+k\}b_i + (1-s)k'e(1-b_i)\over d_i+1}d'
\end{flalign}
Using the above two equations, we can derive the discrimination power as below:
\begin{flalign}
    Z &= kb_i + k'(1-b_i) - \{(s-1)ke+k\}b_i - (1-s)k'e(1-b_i) \\
    &= (1-s)keb+\{1+(s-1)e\}k'(1-b_i) 
\end{flalign}
Given the above equation, we can induce the conditional statement as follows:
\begin{equation}
\begin{gathered}
Z \in
    \begin{cases}
        e+b_i-1-se, & \text{if}\,\, \cos(k',k)=-1\\
        e+b_i-1-se+2eb_i(s-1), & \text{if}\,\, \cos(k',k)=1,
    \end{cases}
\end{gathered}
\end{equation}
Since $\int_{0}^{1}\int_{0}^{1} (e+b_i-1) \,de \, db_i = \left[1-{e^2 + b^2 \over 2}\right]^1_{e,b=0} = 0$, we can induce that the first equation is smaller than 0. Furthermore, if we assume that $s$ is sufficiently small $s \rightarrow 0$, the second equation becomes $\int_{0}^{1}\int_{0}^{1} (e+b_i-1-2eb_i) \, de \, db_i = -1$, which means that edge weight calibration improves the separability of positive GNNs.

\begin{wrapfigure}{r}{0.25\textwidth}
  \centering
    \includegraphics[width=0.2\textwidth]{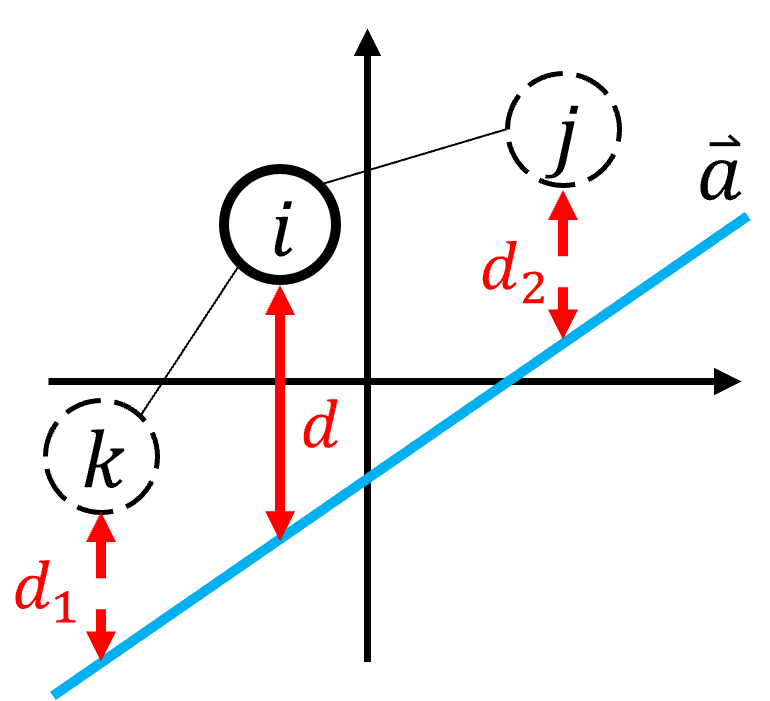}
  \caption{Attention Mechanism}
\end{wrapfigure}

$\\$
\textbf{E-1. Edge weight calibration for GAT \cite{velickovic2017graph}.}
Now, we show that the attention mechanism is not sufficient to capture the similarity of two nodes. Generally, the formulation of attention is defined as below:
\begin{equation}
    a_{ij} = \vec{a}^T[W\vec{h}_i || W\vec{h}_j],
\end{equation}
which is the summation of distances between the attention vector $\vec{a}$ and two nodes $W\vec{h}_i,W\vec{h}_j$ as follows: $a_{ij}=\vec{a}_1^T \cdot W\vec{h}_i + \vec{a}_2^T \cdot W\vec{h}_j$ since attention vector can be decomposed into two vectors $\vec{a}=[\vec{a}_1 || \vec{a}_2]$. Let us assume two neighbors $j$ and $k$ have the same distance from the attention vector. In this case, we can easily infer that the neighbors have identical attention weights since $d+d_1=d+d_2$ and the normalization would remain the same. However, their cosine similarity differs, and we can observe that the attention mechanism may fail to capture this information.

$\\$
\textbf{E-2. Confidence calibration and uncertainty.}

For brevity, assume a star-like subgraph $\mathcal{G}$ consists of an ego node ($i$) and its neighboring nodes ($j$) with homophily ($\mathcal{H}_\mathcal{G}$). Also, let us assume that the label of an ego is 0 ($y_i=0$) under 3 classes $|C|=3$, which is the only node used for training. Firstly, we can easily infer that the prediction of an ego will converge as $\widehat{y}_i=[1, 0, 0]$. Conversely, the neighboring nodes will either be $\widehat{y}_i=[0, 0.5, 0.5]$ or $\widehat{y}_i=[1, 0, 0]$ depending on the sign of connected edges. In this case, the Shannon Entropy can be calculated as $H([1, 0, 0])=0$ and $H([0, 0.5, 0.5])=\log (c-1)=1$. Thus, we can infer that $\mathbb{E}[\mathcal{G}] \approx \mathcal{H}_\mathcal{G} \cdot \log (c-1)$ considering that the neighboring nodes are independent and identically distributed. Now, let's apply the confidence calibration as below:
\begin{equation}
    \mathcal{L}_{conf} = {1 \over N}\sum_{i=1}^N(-\text{max}(\widehat{y}_i)+\text{submax}(\widehat{y}_i))
\end{equation}
Firstly, under perfect homophily $\mathcal{H}_\mathcal{G}=1$, the entropy of all neighbors $H([1, 0, 0])$ that are connected with positive edges will remain the same as 0, where we can induce that confidence calibration is unnecessary $H(\mathbb{E}[\widehat{y}_{cal}])=H(\mathbb{E}[\widehat{y}_s])$ under fully converged condition. However, this condition may not hold under real-world benchmark graphs. Conversely, in case $\mathcal{H}_\mathcal{G} \neq 1$, neighbors with signed edges will be $H([0, 0.5+\lambda, 0.5-\lambda])$ as we explained in Eq. \ref{eq_74}. Since $\alpha$ is proportional to the $\lambda\mathcal{L}_{conf}$ (Eq. \ref{total_loss}), naively assuming that $0 \leq \lambda < 1$, the entropy decreases as $H([0, 0.6, 0.4])=0.971$. As such, the entropy after confidence calibration follows: 
\begin{equation}
    \mathcal{H}_\mathcal{G} \cdot \{ {1 \over c-1}\log ({c-1 \over 1+\lambda}) + {c-2 \over c-1}\log ({c-2 \over 1-\lambda}) \} \leq \mathcal{H}_\mathcal{G} \cdot \{{1 \over c-1}\log ({c-1 \over 1+\lambda}) + {c-2 \over c-1}\log ({c-1 \over 1-\lambda}) \} \leq \mathcal{H}_\mathcal{G} \cdot \log ({c-1 \over 1+\lambda}) 
\end{equation}
Given that the entropy of using signed messages only is $H(\mathbb{E}[\widehat{y}^t_s])$, the expectation of converged entropy ($t \rightarrow \infty$) after confidence calibration ($\mathbb{E}[\widehat{y}^t_{cal}]$) is given by:
\begin{equation}
\lim_{t\to\infty} H(\mathbb{E}[\widehat{y}^t_{cal}]) \approx \underset{\text{signed propagation + calibration}}{\underline{\mathcal{H}_g \cdot \log ({c-1 \over 1+\lambda}) \cdot H(\mathbb{E}[\widehat{y}^t_s])}} \leq \underset{\text{signed propagation only}}{\underline{\mathcal{H}_\mathcal{G} \cdot \log (c-1) \cdot H(\mathbb{E}[\widehat{y}^t_s])}}
\end{equation}
which completes the proof \qedsymbol{}.

In summary, confidence calibration reduces the entropy of positive message-passing. However, it significantly reduces uncertainty in signed propagation as the homophily of the graph decreases. In this context, the frequency of conflict predictions is proportional to the heterophily ratio, which typically exhibits the highest entropy.

$\\$
\textbf{F. Pseudo-code and Time Complexity of Our Algorithm}
$\\$

\textbf{F-1. The pseudo-code of calibrated FAGCN \cite{bo2021beyond} is as below. Instead of FAGCN, any GNNs can be integrated with our method}

\begin{algorithm}[H]
\caption{Pseudo-code: FAGCN \cite{bo2021beyond} with our method}
\label{algo}
\begin{algorithmic}[1]
\Require Adjacency matrix ($A$), initial node features ($X$), node embedding at $l^{th}$ layer ($H^l$), attention weight between two nodes ($a_{ij}$), initialized parameters of FAGCN ($\theta$), edge weight threshold ($\epsilon$), best validation score ($\alpha^*=0$)
\Ensure Parameters with the best validation score ($\theta^*$)
\For{training epochs}
\State Retrieve $l^{th}$ layer's node embedding, $H^l$ 
\State Get node $i's$ embedding, $h^l_i$
\State Get attention weights, $a_{ij}=tanh(g^T\left[h^l_i || h^l_j\right])$ 
\State Normalize cosine similarity for edges in Eq. \ref{method_one}
\State Apply GNNs on $\tilde{a}_{ij}$, $\,\,\widehat{Y}=\sigma(\bar{H}^{(L)})$
\State Compute node classification loss, $\mathcal{L}_{EGNN}$ 
\State Compute calibration loss, $\mathcal{L}_{conf}$ 
\State Get total loss, $\mathcal{L}_{total}$ = $\mathcal{L}_{EGNN}$ + $\lambda \mathcal{L}_{conf}$
\State Update parameters, $\theta'=\theta-\eta{\partial\mathcal{L}_{total} \over \partial\theta}$
\State Using the updated parameters ($\theta'$) and calibrated attention weights ($a_{ij})$, get validation score $\alpha$
\If {$\alpha$ $>$ $\alpha^*$}
\State Save current parameters, $\theta^*=\theta'$
\State Update best validation score, $\alpha^*=\alpha$
\EndIf
\EndFor
\end{algorithmic}
\end{algorithm}

\textbf{F-2. Time complexity of calibrated GCN}

We analyze the computational complexity of calibrated GNN. For brevity, we take vanilla GCN \cite{kipf2016semi} as a base model. Generally, the cost of GCN is known to be proportional to the number of edges and trainable parameters $\mathcal{O}(|\mathcal{E}|\theta_{GCN})$. Here, $\theta_{GCN}$ is comprised of $\mathcal{O}(nz(X)F'+F'C)$ \cite{zhu2020beyond}, where $nz(\cdot)$ represents the non-zero elements of inputs and $F'$ stand for the hidden dimension, and $C$ is the number of classes. 
Additionally, our method employs two types of calibration.
The first one is edge weight calibration. For this, we need to retrieve the node features of each layer and calculate the cosine similarity for all connected nodes $|\mathcal{E}|^2$. Thus, the complexity becomes $\mathcal{O}(|\mathcal{E}|\theta_{GCN}+L|\mathcal{E}|^2$), where $L$ represents the number of entire layers.  
Additionally, confidence calibration takes $n=|\mathcal{V}_{valid} \cup \mathcal{V}_{test}|$ samples as inputs and finds top $k$ samples on each row of $\widehat{Y}$. Thus, their complexity can simply be defined as $\mathcal{O}(n+k)$.
To summarize, the cost of calibrated GCN is $\mathcal{O}(2|\mathcal{E}|\theta_{GCN}+L|\mathcal{E}|^2+n+k)$, which is fairly efficient compared to plane algorithm.

$\\\\$
\textbf{G. Datasets and Implementations Details}


\textbf{G-1. Datasets}

The statistical details of datasets are in Table \ref{dataset}.
(1) \textit{Cora, Citeseer, Pubmed} \cite{kipf2016semi} are citation graphs. where a node corresponds to a paper and edges are citations between them. The labels are the research topics of the papers.    
(2) \textit{Actor} \cite{tang2009social} is a co-occurrence graph of actors in the same movie. The labels are five types of actors. 
(3) \textit{Chameleon, Squirrel} \cite{rozemberczki2019gemsec} are Wikipedia hyperlink networks. Each node is a web page and the edges are hyperlinks. Nodes are categorized into five classes based on monthly traffic. $\\$

\textbf{G-2. General information}

All methods including baselines and ours are implemented upon \textit{PyTorch Geometric}\footnote{\label{code1}https://pytorch-geometric.readthedocs.io/en/latest/modules/nn.html}.
For a fair comparison, we equalize the hidden dimension of the entire methodologies as 64. ReLU with dropout is used for non-linearity and to prevent over-fitting. We employ the log-Softmax as a cross-entropy function.
The learning ratio is set to $1e^{-3}$ and the Adam optimizer is taken with weight decay $5e^{-4}$. For training, 20 nodes per class are randomly chosen and the remaining nodes are divided into two parts for validation and testing, where we followed the settings in \cite{kipf2016semi}. $\\$

\textbf{G-3. Implementation details about baseline methods}

\begin{itemize}
    \item \textbf{GCN} \cite{kipf2016semi} is a first-order approximation of Chebyshev polynomials \cite{defferrard2016convolutional}. For all datasets, we simply take 2 layers of GCN.
    \item \textbf{APPNP} \cite{klicpera2018predict} combines personalized PageRank on GCN. We stack 10 layers and set the teleport probability ($\alpha$) as $\{0.1,0.1,0.1,0.5,0.2,0.3\}$ for Cora, Citeseer, Pubmed, Actor, Chameleon, and Squirrel.
    \item \textbf{GAT} \cite{velickovic2017graph} calculates feature-based attention for edge coefficients. Similar to GCN, we construct 2 layers of GAT. The pair of (hidden dimension, head) is set as (8, 8) for the first layer, while the second layer is (1, \# of classes).
    \item \textbf{GCNII} \cite{chen2020simple} integrates an identity mapping function on APPNP. We set $\alpha=0.5$ and employ nine hidden layers. We increase the weight of identity mapping ($\beta$) that is inversely proportional to the heterophily of the dataset.
    \item \textbf{H$_2$GCN} \cite{zhu2020beyond} suggests the separation of ego and neighbors during aggregation. We refer to the publicly available \textit{source code}\footnote{\label{code2}https://github.com/GemsLab/H2GCN} for implementation. 
    \item \textbf{PTDNet} \cite{luo2021learning} removes disassortative edges before a message-passing. We also utilize the open \textit{source code}\footnote{\label{code3}https://github.com/flyingdoog/PTDNet} and apply confidence calibration.
    \item \textbf{GPRGNN} \cite{chien2020adaptive} generalized the personalized PageRank to deal with heterophily and over-smoothing. Referring to the open source \textit{code}\footnote{\label{code4}https://github.com/jianhao2016/GPRGNN}, we tune the hyper-parameters based on the best validation score for each dataset.
    \item \textbf{FAGCN} \cite{bo2021beyond} determines the sign of edges using the node features. We implement the algorithm based on the \textit{sources}\footnote{\label{code5}https://github.com/bdy9527/FAGCN} and tune the hyper-parameters concerning their accuracy.
    \item \textbf{GGCN} \cite{yan2021two} proposes the scaling of degrees and the separation of positive/negative adjacency matrices. We simply take the publicly available \textit{code}\footnote{\label{code6}https://github.com/Yujun-Yan/Heterophily\text{\_}and\text{\_}oversmoothing} for evaluation.
\end{itemize}

\begin{figure*}[t]
    \includegraphics[width=\textwidth]{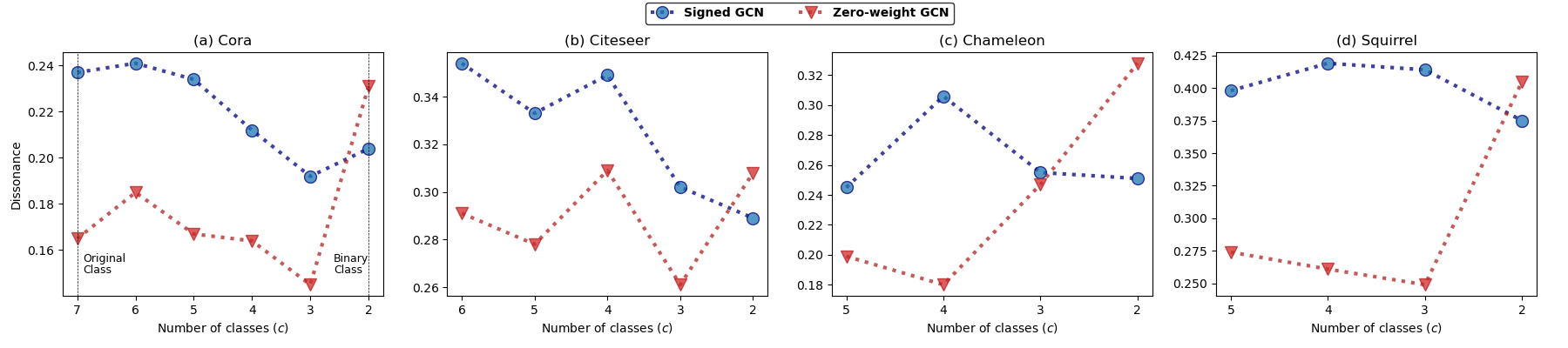}
    \caption{By differentiating the number of classes, we compare the dissonance of signed GCN and zero-weight GCN using two graph variants. The leftmost value on the x-axis represents the original class, and the rightmost value represents the binary class}
  \label{case_study}
\end{figure*}

$\\\\$
\textbf{H. Further Evaluation}
$\\$

\textbf{H-1. Case study}

Through theoretical analyses, we confirm that signed messages increase the uncertainty in multi-class graphs. They have shown to be effective when $k'$ gets closer to $-k$ (Eq. \ref{class_prob}), but this probability is inversely proportional to the number of classes \textit{c}. To further analyze this phenomenon, we compare the dissonance of two variants of GCN (signed GCN and zero-weight GCN) by decrement of the number of classes (\textit{c}). Specifically, if the original data contains seven classes (e.g., Cora), we remove all the nodes that belong to the rightmost class to generate a graph with six labels. The results are illustrated in Figure \ref{case_study}. As can be seen, zero-weight GCN (red) tends to have lower dissonance under multiple classes. However, under binary classes (\textit{c=2}), signed GCN (blue) shows lower uncertainty with the aid of ego-neighbor separation. In the binary case, zero-weight GCN only utilizes homophilous neighbors and fails to generalize under this condition.
$\\$

\begin{table*}[ht]
\caption{Node classification accuracy (\%) on nine benchmark graphs. The gray-colored cell means the top 3 and bold with ($^\ast$) indicates the best performances. Symbol $\ddagger$ means that calibration is applied to the baseline method}
\label{perf_2}
\centering
\begin{center}
\begin{adjustbox}{width=\textwidth}
\begin{tabular}{@{}lllllllllll}
&  & & &              \\ 
\Xhline{2\arrayrulewidth}
        & Datasets                       & Cora  & Citeseer & Pubmed & Actor & Chameleon & Squirrel & Cornell & Texas & Wisconsin \\ 
\Xhline{2\arrayrulewidth}
        & $\mathcal{H}_g$ (Eq. \ref{global_homo}) & 0.81 & 0.74 & 0.8 & 0.22 & 0.23 & 0.22 & 0.11 & 0.06 & 0.16 \\
\Xhline{2\arrayrulewidth}
                        & MLP              & 53.2 $_{\,\pm\,0.5 }$ & 53.7 $_{\,\pm\,1.7 }$ & 69.7 $_{\,\pm\,0.4 }$ & \cellcolor{Gray}27.9 $_{\,\pm\,1.1 }$ & 41.2 $_{\,\pm\,1.8 }$ & 26.5 $_{\,\pm\,0.6 }$ & 60.1 $_{\,\pm\,1.2 }$ & 65.8 $_{\,\pm\,5.0 }$ & \cellcolor{Gray}\textbf{73.5} $^*_{\,\pm\,5.4 }$\\
                        & GCN                & 79.0 $_{\,\pm\,0.6 }$ & 67.5 $_{\,\pm\,0.8 }$  & 77.6 $_{\,\pm\,0.2 }$  & 20.2 $_{\,\pm\,0.4 }$   & 49.3 $_{\,\pm\,0.5 }$  & 30.7 $_{\,\pm\,0.7 }$  & 39.4 $_{\,\pm\,4.3 }$ & 47.6 $_{\,\pm\,0.7 }$ & 40.5 $_{\,\pm\,1.9 }$ \\
                        & \textbf{GCN}$^\ddagger$       & 81.0 $_{\,\pm\,0.9 }$  & 71.3 $_{\,\pm\,1.2 }$ & 77.8 $_{\,\pm\,0.4 }$ & 21.7 $_{\,\pm\,0.6 }$ & 49.4 $_{\,\pm\,0.6 }$ & 31.5 $_{\,\pm\,0.6 }$ & 39.6 $_{\,\pm\,3.1 }$ & 48.0 $_{\,\pm\,0.6 }$ & 40.5 $_{\,\pm\,1.4 }$\\
                        & GAT             & 80.1 $_{\,\pm\,0.6 }$ & 68.0 $_{\,\pm\,0.7 }$ & 78.0 $_{\,\pm\,0.4 }$ & 22.5 $_{\,\pm\,0.3 }$ & 47.9 $_{\,\pm\,0.8 }$ & 30.8 $_{\,\pm\,0.9 }$ & 42.1 $_{\,\pm\,3.1 }$ & 49.2 $_{\,\pm\,4.4 }$ & 45.8 $_{\,\pm\,5.3 }$ \\
                        & \textbf{GAT}$^\ddagger$             & 81.4 $_{\,\pm\,0.4 }$ & 72.2 $_{\,\pm\,0.6 }$ & 78.3 $_{\,\pm\,0.3 }$ & 23.2 $_{\,\pm\,1.8 }$ & 49.2 $_{\,\pm\,0.4 }$ & 30.3 $_{\,\pm\,0.8 }$ & 42.5 $_{\,\pm\,2.5 }$ & 49.3 $_{\,\pm\,4.1 }$ & 47.1 $_{\,\pm\,5.0 }$ \\
                        & GATv2             & 79.5 $_{\,\pm\,0.5 }$ & 67.4 $_{\,\pm\,0.6 }$ & 76.2 $_{\,\pm\,0.5 }$ & 22.1 $_{\,\pm\,2.0 }$ & 48.3 $_{\,\pm\,0.4 }$ & 28.9 $_{\,\pm\,1.2 }$ & 38.0 $_{\,\pm\,3.8 }$ & 52.5 $_{\,\pm\,1.7 }$ & 41.7 $_{\,\pm\,5.1 }$ \\
                        & GIN              & 77.3 $_{\,\pm\,0.8 }$ & 66.1 $_{\,\pm\,0.6 }$ & 77.1 $_{\,\pm\,0.7 }$ & 24.6 $_{\,\pm\,0.8 }$ & 49.1 $_{\,\pm\,0.7 }$ & 28.4 $_{\,\pm\,2.2 }$ & 42.9 $_{\,\pm\,4.6 }$ & 53.5 $_{\,\pm\,3.0 }$ & 38.2 $_{\,\pm\,1.5 }$\\
                        & APPNP                 & 81.3 $_{\,\pm\,0.5 }$ & 68.9 $_{\,\pm\,0.3 }$ & 79.0 $_{\,\pm\,0.3 }$ & 23.8 $_{\,\pm\,0.3 }$ & 48.0 $_{\,\pm\,0.7 }$ & 30.4 $_{\,\pm\,0.6 }$ & 49.8 $_{\,\pm\,3.6 }$ & 56.1 $_{\,\pm\,0.2 }$ & 45.7 $_{\,\pm\,1.7 }$\\
                        & GCNII           & 81.1 $_{\,\pm\,0.7 }$ & 68.5 $_{\,\pm\,1.4 }$ & 78.5 $_{\,\pm\,0.4 }$ & 25.9 $_{\,\pm\,1.2 }$ & 48.1 $_{\,\pm\,0.7 }$ & 29.1 $_{\,\pm\,0.9 }$ & 62.5 $_{\,\pm\,0.5 }$ & \cellcolor{Gray}69.3 $_{\,\pm\,2.1 }$ & \cellcolor{Gray}63.2 $_{\,\pm\,3.0 }$ \\
                        & H$_2$GCN        & 80.6 $_{\,\pm\,0.6 }$ & 68.2 $_{\,\pm\,0.7 }$ & 78.5 $_{\,\pm\,0.3 }$ & 25.6 $_{\,\pm\,1.0 }$ & 47.3 $_{\,\pm\,0.8 }$ & 31.3 $_{\,\pm\,0.7 }$ & 59.8 $_{\,\pm\,3.7 }$ & 66.3 $_{\,\pm\,4.6 }$ & 61.5 $_{\,\pm\,4.4 }$ \\
                        & Ortho-GCN        & 80.6 $_{\,\pm\,0.4 }$ & 69.5 $_{\,\pm\,0.3 }$ & 76.9 $_{\,\pm\,0.3 }$ & 21.4 $_{\,\pm\,1.6 }$ & 46.7 $_{\,\pm\,0.5 }$ & 31.3 $_{\,\pm\,0.6 }$ & 45.4 $_{\,\pm\,4.7 }$ & 53.1 $_{\,\pm\,3.9 }$ & 46.6 $_{\,\pm\,5.8 }$ \\
                        & P-reg       & 80.0 $_{\,\pm\,0.8 }$ & 69.2 $_{\,\pm\,0.7 }$ & 77.4 $_{\,\pm\,0.4 }$ & 20.9 $_{\,\pm\,0.5 }$ & 49.1 $_{\,\pm\,0.1 }$ & \cellcolor{Gray}\textbf{33.6$^*$} $_{\,\pm\,0.4 }$ & 44.9 $_{\,\pm\,3.1 }$ & 58.5 $_{\,\pm\,4.2 }$ & 53.7 $_{\,\pm\,2.6 }$ \\
                        & ACM-GCN & 80.2 $_{\,\pm\,0.8 }$ & 68.3 $_{\,\pm\,1.1 }$ & 78.1 $_{\,\pm\,0.5 }$ & 24.9 $_{\,\pm\,2.0 }$ & 49.5 $_{\,\pm\,0.7 }$ & 31.6 $_{\,\pm\,0.4 }$ & 55.6 $_{\,\pm\,3.3 }$ & 58.9 $_{\,\pm\,2.6 }$ & 61.3 $_{\,\pm\,0.5 }$ \\
                        & HOG-GCN & 79.7 $_{\,\pm\,0.4 }$ & 68.2 $_{\,\pm\,0.6 }$ & 78.0 $_{\,\pm\,0.2 }$ & 21.5 $_{\,\pm\,0.5 }$ & 47.7 $_{\,\pm\,0.5 }$ & 30.1 $_{\,\pm\,0.4 }$ & 61.8 $_{\,\pm\,0.8 }$ & 60.4 $_{\,\pm\,0.6 }$ & 62.7 $_{\,\pm\,0.3 }$ \\
                        & JacobiConv & 81.9 $_{\,\pm\,0.6 }$ & 69.6 $_{\,\pm\,0.8 }$ & 78.5 $_{\,\pm\,0.4 }$ & 25.7 $_{\,\pm\,1.2 }$ & \cellcolor{Gray}\textbf{52.8$^*$ $_{\,\pm\,0.9 }$} & 32.0 $_{\,\pm\,0.6 }$ & 55.3 $_{\,\pm\,3.4 }$ & 57.7 $_{\,\pm\,3.6 }$ & 53.4 $_{\,\pm\,1.6 }$ \\
                        & GloGNN & 82.4 $_{\,\pm\,0.3 }$ & 70.3 $_{\,\pm\,0.5 }$ & \cellcolor{Gray}79.3 $_{\,\pm\,0.2 }$ & 26.6 $_{\,\pm\,0.7 }$ & 48.2 $_{\,\pm\,0.3 }$ & 28.8 $_{\,\pm\,0.8 }$ & 56.8 $_{\,\pm\,1.1 }$ & 63.0 $_{\,\pm\,2.9 }$ & 59.5 $_{\,\pm\,1.3 }$ \\
                        & TED-GCN   & 81.8 $_{\,\pm\,0.9 }$  & 71.4 $_{\,\pm\,0.6 }$ & 78.6 $_{\,\pm\,0.3 }$  & 26.0 $_{\,\pm\,0.9 }$  & 50.4 $_{\,\pm\,1.2 }$  & \cellcolor{Gray}33.0 $_{\,\pm\,0.9 }$ & 57.3 $_{\,\pm\,1.2 }$ & 60.5 $_{\,\pm\,1.6 }$ & 49.2 $_{\,\pm\,2.8 }$ \\
                        & PCNet  & 81.5 $_{\,\pm\,0.8 }$  & 71.2 $_{\,\pm\,1.2 }$  & 78.8 $_{\,\pm\,0.3 }$  & 26.4 $_{\,\pm\,0.8 }$  & 48.1 $_{\,\pm\,1.7 }$  & 31.4 $_{\,\pm\,1.6 }$  & 55.0 $_{\,\pm\,1.7 }$  & 59.2 $_{\,\pm\,1.6 }$  & 54.4 $_{\,\pm\,1.1 }$ \\
\Xhline{2\arrayrulewidth}
                        & PTDNet          & 81.2 $_{\,\pm\,0.9 }$  & 69.5 $_{\,\pm\,1.2 }$  & 78.8 $_{\,\pm\,0.5 }$  & 21.5 $_{\,\pm\,0.6 }$  & 50.6 $_{\,\pm\,0.9 }$  & 32.1 $_{\,\pm\,0.7 }$ & \cellcolor{Gray}63.3 $_{\,\pm\,1.1 }$ & 59.8 $_{\,\pm\,9.2 }$ & 53.1 $_{\,\pm\,5.6 }$ \\
                        & \textbf{PTDNet}$^\ddagger$     & 81.9 $_{\,\pm\,0.6 }$  & 71.1 $_{\,\pm\,0.8 }$  & 79.0 $_{\,\pm\,0.2 }$  & 22.7 $_{\,\pm\,0.6 }$  & \cellcolor{Gray}50.9 $_{\,\pm\,0.3 }$  & \cellcolor{Gray}32.3 $_{\,\pm\,0.5 }$ & \cellcolor{Gray}64.4 $_{\,\pm\,3.3 }$ & 62.4 $_{\,\pm\,6.0 }$ & 55.7 $_{\,\pm\,3.7 }$\\
\Xhline{2\arrayrulewidth}
                        & GPR-GNN & 82.2 $_{\,\pm\,0.4 }$ & 70.4 $_{\,\pm\,0.8 }$ & 79.1 $_{\,\pm\,0.1 }$ & 25.4 $_{\,\pm\,0.5 }$ & 49.1 $_{\,\pm\,0.7 }$ & 30.5 $_{\,\pm\,0.6 }$ & 57.1 $_{\,\pm\,1.6 }$ & 61.2 $_{\,\pm\,0.9 }$ & 62.4 $_{\,\pm\,1.2 }$ \\
                        & \textbf{GPRGNN}$^\ddagger$ & \cellcolor{Gray}\textbf{84.7 $^*_{\,\pm\,0.2 }$}  & \cellcolor{Gray}73.3 $_{\,\pm\,0.5 }$  & \cellcolor{Gray}\textbf{80.2} $^*_{\,\pm\,0.2 }$  & \cellcolor{Gray}\textbf{28.1} $^*_{\,\pm\,1.3 }$  & \cellcolor{Gray}51.0 $_{\,\pm\,0.4 }$ & 31.8 $_{\,\pm\,0.4 }$ & 61.1 $_{\,\pm\,2.8 }$ & 63.0 $_{\,\pm\,1.4 }$ & \cellcolor{Gray}63.2 $_{\,\pm\,1.5}$\\
                        & FAGCN        & 81.9 $_{\,\pm\,0.5 }$  & 70.8 $_{\,\pm\,0.6 }$  & 79.0 $_{\,\pm\,0.5 }$  & 25.2 $_{\,\pm\,0.8 }$  & 46.5 $_{\,\pm\,1.1 }$  & 30.4 $_{\,\pm\,0.4 }$ & 55.6 $_{\,\pm\,3.0 }$ & 50.1 $_{\,\pm\,3.9 }$ & 57.8 $_{\,\pm\,6.5 }$ \\
                        & \textbf{FAGCN}$^\ddagger$ & \cellcolor{Gray}84.1 $_{\,\pm\,0.4 }$  & \cellcolor{Gray}\textbf{73.8 $^*_{\,\pm\,0.5 }$}  & \cellcolor{Gray}79.7 $_{\,\pm\,0.2 }$  & \cellcolor{Gray}27.6 $_{\,\pm\,0.5 }$  & 48.8 $_{\,\pm\,0.7 }$  & 31.3 $_{\,\pm\,0.5 }$ & 60.5 $_{\,\pm\,2.9 }$ & 55.4 $_{\,\pm\,2.6 }$ & 59.6 $_{\,\pm\,4.6 }$ \\
                        & GGCN & 81.0 $_{\,\pm\,1.2 }$  & 70.7 $_{\,\pm\,1.6 }$  & 78.2 $_{\,\pm\,0.4 }$  & 22.5 $_{\,\pm\,0.5 }$  & 48.5 $_{\,\pm\,0.7 }$  & 30.2 $_{\,\pm\,0.6 }$ & 62.8 $_{\,\pm\,4.0 }$ & \cellcolor{Gray}66.4 $_{\,\pm\,1.7 }$ & 61.7 $_{\,\pm\,3.2 }$\\
                        & \textbf{GGCN}$^\ddagger$ & \cellcolor{Gray}83.9 $_{\,\pm\,0.8 }$  & \cellcolor{Gray}73.0 $_{\,\pm\,0.4 }$  & 78.9 $_{\,\pm\,0.3 }$  & 24.6 $_{\,\pm\,0.4 }$  & 50.0 $_{\,\pm\,0.4 }$  & 31.1 $_{\,\pm\,0.6 }$ & \cellcolor{Gray}\textbf{66.7} $^*_{\,\pm\,2.2 }$ & \cellcolor{Gray}\textbf{69.9} $^*_{\,\pm\,0.6 }$ & 62.6 $_{\,\pm\,1.5 }$\\
\Xhline{2\arrayrulewidth}
\end{tabular}
\end{adjustbox}
\end{center}
\end{table*}

\textbf{H-2. Node classification accuracy of other state-of-the-art methods}

In Table \ref{perf_2}, we measure the node classification accuracy by adding three datasets and various state-of-the-art baselines. The WebKB graphs \textit{Cornell, Texas, Wisconsin\footnote{http://www.cs.cmu.edu/~webkb/}} are comprised of the web pages of computer science departments in various universities. We briefly introduce the details of newly added methods and their implementation as follows:
\begin{itemize}
    \item \textbf{P-reg} \cite{yang2021rethinking} ensembles a regularization term to provide additional information that training nodes might not capture.(\textit{source code}\footnote{https://github.com/yang-han/P-reg})
    \item \textbf{ACM-GCN} \cite{luan2022revisiting} suggests a local diversification operation through the adaptive channel mixing algorithm. (\textit{source code}\footnote{https://github.com/SitaoLuan/ACM-GNN}) 
    \item \textbf{HOG-GCN} \cite{wang2022powerfulb} adaptively controls the propagation mechanism by measuring the homophily degrees between two nodes. (\textit{source code}\footnote{https://github.com/hedongxiao-tju/HOG-GCN})
    \item \textbf{JacobiConv} \cite{wang2022powerfula} studies the expressive power of spectral GNN and establishes a connection with the graph isomorphism testing. (\textit{source code}\footnote{https://github.com/GraphPKU/JacobiConv})
    \item \textbf{GloGNN} \cite{li2022finding} receives information from global nodes, which can accelerate neighborhood aggregation. (\textit{source code}\footnote{https://github.com/RecklessRonan/GloGNN})
    \item \textbf{TED-GCN} \cite{yan2024trainable} redefines GCN’s depth $L$ as a trainable parameter, which can control its signal processing capability to model both homophily/heterophily graphs.
    \item \textbf{PCNet} \cite{li2024pc} proposes a two-fold filtering mechanism to extract homophily in heterophilic graphs \textit{(source code)}\footnote{https://github.com/uestclbh/PC-Conv}.
\end{itemize}

In the above table, we can see that signed GNNs (GPRGNN, FAGCN, and GGCN) with calibration generally outperform the baseline methods.
As mentioned before, signed methods achieve inferior performance on the Chameleon and Squirrel datasets, where many nodes share the same neighboring nodes. Under these conditions, PTDNet outperforms the signed methods, which justifies our analysis.
However, in another graph, calibrated($\ddagger$) GNNs still achieve competitive results. Especially, for the WebKB datasets, the average improvement of signed GNNs through calibration is 3.7\%, 7.5\%, and 4.4\%, respectively.
This result demonstrates that calibration can significantly enhance signed propagation, leading to state-of-the-art performance on real-world benchmark graph datasets.

\textbf{Remark.} Note that the edge classification error ratio ($e$) in Equation \ref{class_prob} is proportional to the performance of the GNN used as the base model. Thus, if the value of $e$ for a model is high enough, blocking information might help improve accuracy rather than the signed propagation.

\end{appendices}

\end{document}